\documentclass{article}

\usepackage{xcolor}
\usepackage{array}

\usepackage{PRIMEarxiv}
\usepackage{subcaption}

\usepackage[utf8]{inputenc} 
\usepackage[T1]{fontenc}    
\usepackage{hyperref}       
\usepackage{url}            
\usepackage{booktabs}       
\usepackage{amsfonts}       
\usepackage{nicefrac}       
\usepackage{microtype}      
\usepackage{lipsum}
\usepackage{fancyhdr}       
\usepackage{graphicx}       
\graphicspath{{./}{figures/}{images/}}     
\usepackage{amsmath}
\usepackage{tabularx}

\title{Comprehensive Evaluation of Monocular and Multi-view 3D Human Pose Estimation Models for Fitness Exercise Analysis}

\author{
  Néstor Ortega \\
  Centro de Tecnologías de la Imagen (CTIM)\\
  Instituto Universitario de Cibernética, Empresas y Sociedad (IUCES)\\
  University of Las Palmas de Gran Canaria \\
  35017 Las Palmas de Gran Canaria, Spain\\
  \texttt{nestor.ortega101@ulpgc.es} \\
  ORCID: \href{https://orcid.org/0009-0001-1708-0428}{0009-0001-1708-0428} \\
\And
  Agustín Salgado \\
  Centro de Tecnologías de la Imagen (CTIM)\\
  Instituto Universitario de Cibernética, Empresas y Sociedad (IUCES)\\
  University of Las Palmas de Gran Canaria \\
  35017 Las Palmas de Gran Canaria, Spain\\
  \texttt{agustin.salgado@ulpgc.es} \\
  ORCID: \href{https://orcid.org/0000-0002-6142-3432}{0000-0002-6142-3432} \\
\And
  Javier Sánchez \\
  Centro de Tecnologías de la Imagen (CTIM)\\
  Instituto Universitario de Cibernética, Empresas y Sociedad (IUCES)\\
  University of Las Palmas de Gran Canaria \\
  35017 Las Palmas de Gran Canaria, Spain\\
  \texttt{jsanchez@ulpgc.es} \\
  ORCID: \href{https://orcid.org/0000-0001-8514-4350}{0000-0001-8514-4350} \\
}

\begin{document}
\maketitle

\begin{abstract}
Three-dimensional human pose estimation has become a fundamental component of computer vision systems for sports analytics, rehabilitation, and motion understanding. Although recent deep learning approaches have achieved remarkable accuracy on standard benchmarks, their performance under realistic sports scenarios remains insufficiently studied. Furthermore, the trade-offs between recent monocular and multi-view approaches, as well as the benefits of post-processing techniques for improving temporal consistency, have not been systematically evaluated on fitness-oriented datasets. This paper evaluates recent state-of-the-art 3D human pose estimation methods on Fit3D, a dataset of synchronized multi-view recordings of a wide variety of fitness exercises with accurate ground-truth annotations. We study two recent monocular lifting approaches, TCPFormer and PoseMamba-L, and a multi-view model, MVGFormer, under a unified evaluation protocol. To investigate the influence of the input data, monocular methods are evaluated using ground-truth keypoints and detections generated by recent 2D pose estimators, RTMPose and YOLO-Pose. Additionally, we analyze the effectiveness of three temporal filters for reducing prediction jitter while preserving motion dynamics. Experimental results provide a detailed comparison of accuracy, temporal consistency, and robustness to imperfect 2D detections across the evaluated methods. Our analysis highlights the impact of detector quality on lifting-based approaches, compares transformer- and state-space-based temporal architectures, and discusses the practical trade-offs between monocular and multi-view systems for fitness motion analysis. Our results show that 2D detection quality, temporal modeling, and post-processing must be considered jointly in pose estimation pipelines for exercise analysis.
\end{abstract}


\section{Introduction}
Three-dimensional human pose estimation~\cite{wang2021deep,neupane2025survey} aims to recover the spatial configuration of the human body from one or more images or video sequences. Over the last decade, it has become one of the most active research areas in computer vision due to its wide range of applications, including sports performance analysis~\cite{ucar2026sports}, physical rehabilitation~\cite{rehab2462025}, virtual and augmented reality, human--computer interaction, and computer animation. Accurate reconstruction of body joints in three dimensions enables detailed analysis of human movement that cannot be obtained from conventional 2D image representations. 

Recent advances in deep learning have substantially improved the accuracy of 3D human pose estimation. Monocular methods have evolved from relatively simple frame-wise 2D-to-3D lifting approaches~\cite{martinez2017simple} to sophisticated temporal architectures capable of exploiting motion information over long video sequences~\cite{zhang2022mixste}. In particular, Transformer-based models have demonstrated remarkable performance by modeling long-range dependencies between body joints and temporal observations. More recently, state-space models~\cite{huang2025posemamba} have emerged as an efficient alternative, providing comparable accuracy while reducing computational complexity. Multi-view approaches~\cite{tu2020voxelpose,neupane2025survey} provide highly accurate 3D pose estimates when synchronized cameras and calibration parameters are available, as they can explicitly exploit geometric constraints to reduce depth ambiguities. 

Despite these advances, several challenges remain before these methods can be reliably deployed in practical applications. One important limitation is that many monocular approaches rely on a two-stage pipeline in which 2D keypoints are first detected and subsequently lifted into 3D. Consequently, the final reconstruction accuracy depends on the quality of the 2D detections. Although modern pose detectors perform well, they can still produce errors under occlusions, motion blur, and other challenging visual conditions. These errors can propagate through lifting and substantially degrade the resulting 3D pose~\cite{hoang2024robustness}. Therefore, the robustness of recent 3D pose lifters to realistic 2D detection errors remains an important practical consideration.

Another challenge concerns temporal consistency. While current methods typically optimize spatial reconstruction accuracy, many applications require temporally consistent pose estimates. Small frame-to-frame fluctuations, even with low spatial errors, may significantly affect downstream tasks such as biomechanical analysis, exercise assessment, motion segmentation, or movement quality evaluation. This is particularly relevant for multi-view methods, which typically reconstruct each frame independently and whose temporal behavior has received little attention. Temporal smoothing techniques are often used as a post-processing step~\cite{zeng2022smoothnet}, yet their effectiveness across different 3D pose estimation models remains underexplored.

Moreover, most methods are trained and compared on laboratory benchmarks such as Human3.6M~\cite{ionescu2014human36m}, whose everyday actions only partially reflect fitness exercises, with their non-upright poses, self-occlusions, and large joint excursions. In this work, we evaluate recent state-of-the-art methods on the Fit3D dataset~\cite{fit3d2024}, which provides synchronized multi-view recordings of diverse fitness exercises with accurate 3D ground truth. All models are used with their released checkpoints, without retraining, so the evaluation also measures their cross-dataset generalization. Our study focuses on three representative architectures covering both monocular and multi-view paradigms. Specifically, we evaluate TCPFormer~\cite{liu2025tcpformer}, a transformer-based temporal lifting model, PoseMamba~\cite{huang2025posemamba}, a recent state-space-based architecture designed for efficient temporal modeling, and MVGFormer~\cite{liao2024mvgformer}, a multi-view transformer that integrates geometric reasoning with deep feature learning. 

Lifting methods are usually compared on Human3.6M, and the effect of recent detectors such as RTMPose~\cite{jiang2024rtmpose} and YOLO-Pose~\cite{maji2022yolopose} has not been quantified on exercise sequences. We therefore investigate several complementary aspects that directly affect the practical deployment of 3D pose estimation systems. First, we analyze how the quality of the input 2D detections influences the performance of modern lifting methods by comparing ground-truth keypoints with the predictions of these two detectors. Second, we compare Transformer-based and state-space-based temporal representations to understand their respective strengths and limitations for complex fitness motions. Third, we evaluate the advantages and limitations of monocular and multi-view approaches. Finally, we investigate whether lightweight temporal filters, such as the One Euro~\cite{casiez2012oneeuro}, Savitzky--Golay~\cite{savitzky1964smoothing}, and median filters, improve the temporal consistency of reconstructed poses without degrading spatial accuracy.

The contributions of this work can be summarized as follows: i) our study presents a unified experimental evaluation of recent monocular and multi-view 3D human pose estimation models on the Fit3D benchmark using a common evaluation protocol; ii) we quantify the influence of different 2D pose detectors on the performance of modern lifting-based methods, providing insight into the robustness of current monocular pipelines; and iii) we analyze the effectiveness of temporal post-processing techniques for improving the temporal consistency of 3D pose predictions in fitness motion sequences.

Our results show that replacing ground-truth 2D keypoints with RTMPose detections increases the PA-MPJPE of the monocular models from 37--46\,mm to about 71\,mm, that MVGFormer obtains the lowest spatial error but an MPJVE about nine times higher than the monocular models, and that median filtering reduces this temporal error by 84\% while also improving its spatial accuracy. 

This article is organized as follows: Section~\ref{se:related_work} reviews related work on 3D pose benchmarks, 2D pose detection, monocular and multi-view 3D pose estimation, and temporal post-processing. Section~\ref{se:methods} describes the evaluation framework, including the evaluated models, the input representations, and the temporal filters. Section~\ref{se:experiments} presents the dataset, the evaluation metrics, the implementation details, and the experimental results. Finally, Section~\ref{se:discussion} discusses the findings and limitations of this work, and Section~\ref{se:conclusion} summarizes the main conclusions and outlines future research directions.

\section{Related work}
\label{se:related_work}

Three-dimensional human pose estimation is widely used in sports and exercise analysis~\cite{ucar2026sports}. This section reviews the available benchmarks, the role of the 2D detector in two-stage pipelines, the evolution of temporal modeling in monocular lifting, multi-view reconstruction, and post-processing.

\subsection{3D pose benchmarks and exercise analysis}
\label{sec:rw_benchmarks}

In the fitness and rehabilitation domains, pose and motion analysis is used to count repetitions and quantify exercise execution~\cite{stromback2020mmfit}, assess the quality of an action~\cite{parmar2019aqa}, provide automatic corrective feedback~\cite{fit3d2024}, and monitor physical rehabilitation~\cite{rehab2462025}. These applications require not only accurate poses at each frame but also temporally consistent sequences.

Most 3D pose estimation methods, however, are developed and compared on a small number of laboratory benchmarks, such as Human3.6M~\cite{ionescu2014human36m} and HumanEva~\cite{sigal2010humaneva}, which contain everyday actions. MPI-INF-3DHP~\cite{mehta2017mpiinf3dhp} involves outdoor scenes and more varied poses, and CMU Panoptic~\cite{joo2015panoptic} provides dense multi-view recordings of social interactions. Datasets with more dynamic motions exist, such as AIST++~\cite{li2021aist} for dance, but fitness exercises involve large joint excursions, frequent self-occlusions, and body configurations, such as lying or kneeling poses, that are rare in these benchmarks. Fit3D~\cite{fit3d2024} was collected specifically for this domain and provides multi-view video and motion-capture ground truth for a wide range of exercises. Because models are usually trained on a single benchmark, their accuracy drops substantially across datasets. Our study evaluates models on Fit3D without retraining, which directly measures this generalization in a realistic exercise-analysis setting.

\subsection{2D pose detection}
\label{sec:rw_2d}

Most monocular 3D methods follow a two-stage pipeline in which 2D keypoints are first detected and then lifted to 3D. 2D detectors are typically trained on COCO~\cite{lin2014coco}, which defines the widely used 17-keypoint format. Bottom-up methods, such as OpenPose~\cite{cao2017openpose}, detect all joints in the image and then group them into individuals, whereas top-down methods first detect each person and then estimate the pose inside the bounding box. The latter approach, followed by HRNet~\cite{wang2020hrnet} and ViTPose~\cite{xu2022vitpose}, generally achieves higher accuracy. Recent detectors trade off accuracy and speed: RTMPose~\cite{jiang2024rtmpose} is a top-down model designed for real-time deployment, while one-stage methods such as YOLO-Pose~\cite{maji2022yolopose} predict people and keypoints in a single pass.

The quality of these detections directly affects the lifted 3D pose. Martinez et al.~\cite{martinez2017simple} showed that a simple network lifting ground-truth 2D keypoints achieves much lower errors than the same network fed with detected keypoints, indicating that a large part of the 3D error originates from the 2D stage. Since then, lifting methods have usually been reported on Human3.6M with both ground-truth inputs and 2D detections~\cite{pavllo2019videopose3d,zhu2023motionbert}. However, the effect of recent detectors such as RTMPose or YOLO-Pose on recent lifting architectures has not been systematically quantified in exercise sequences, which motivates the comparison of 2D sources performed in this work.

\subsection{Monocular 3D pose estimation and temporal modeling}
Monocular methods either regress 3D joints directly from images~\cite{pavlakos2017coarse,sun2018integral} or lift 2D keypoints to 3D~\cite{martinez2017simple}. The latter has become the dominant approach because 2D detectors can be trained on large datasets and the lifter only needs pairs of 2D and 3D poses.

Temporal information has traditionally been exploited with recurrent networks~\cite{hossain2018lstm}, dilated temporal convolutions~\cite{pavllo2019videopose3d} and spatio-temporal graph convolutions~\cite{cai2019graph}. Following the success of the Transformer~\cite{vaswani2017attention}, PoseFormer~\cite{zheng2021poseformer} introduced separate spatial and temporal attention, MHFormer~\cite{li2022mhformer} generated multiple pose hypotheses, and MixSTE~\cite{zhang2022mixste} combined spatial and temporal attention to predict the whole sequence at once. MotionBERT~\cite{zhu2023motionbert} added large-scale pretraining on heterogeneous motion data, MotionAGFormer~\cite{li2024motionagformer} combined attention with graph convolutions, and KTPFormer~\cite{peng2024ktpformer} incorporated kinematic and trajectory priors. Other works model the 3D ambiguity with diffusion models~\cite{shan2023d3dp}. TCPFormer~\cite{liu2025tcpformer}, one of the models used in this work, introduces an implicit pose proxy that summarizes the sequence and helps learn temporal correlations.

A limitation of attention is that its cost grows quadratically with the number of frames. State-space models, particularly Mamba~\cite{gu2024mamba} architectures, offer an alternative with linear complexity and a strong ability to model long sequences. PoseMamba~\cite{huang2025posemamba}, the second monocular model evaluated here, applies bidirectional state-space scans over joints and frames and reaches accuracy comparable to Transformer-based models with a fraction of their parameters and computation. Both families have been compared almost exclusively on Human3.6M and MPI-INF-3DHP. It remains unclear how they behave on unseen, more dynamic motions and whether their temporal modeling yields consistent trajectories.

\subsection{Multi-view 3D pose estimation}
\label{sec:rw_multiview}

When several calibrated cameras are available, the depth ambiguity of monocular estimation can be resolved geometrically. The classical approach detects 2D keypoints in every view and triangulates them~\cite{hartley2004multiple}, which is simple but sensitive to detection errors and occlusions in individual views. Learning-based methods combine the information of various views before or during reconstruction: learnable triangulation~\cite{iskakov2019learnable} weights each view according to its confidence or aggregates features in a 3D volume, cross-view fusion~\cite{qiu2019crossview} and epipolar transformers~\cite{he2020epipolar} use epipolar geometry to improve the 2D features of each view, and VoxelPose~\cite{tu2020voxelpose} and Faster VoxelPose~\cite{ye2022fastervoxelpose} reason directly in a discretized 3D space. MvP~\cite{wang2021mvp} replaced voxels with Transformer queries that directly regress multi-person 3D poses. These methods are, however, usually trained for a specific camera arrangement, and their accuracy decreases when the number or position of cameras changes. MVGFormer~\cite{liao2024mvgformer} addresses this problem by alternating learning-free geometric modules based on differentiable triangulation with learned appearance modules, which yields strong generalization to new camera setups. This property makes it well suited to our cross-dataset evaluation. Like most multi-view approaches, it processes each time instant independently, so its temporal behavior has received little attention.

\subsection{Temporal consistency and post-processing}
\label{sec:rw_temporal}

Spatial accuracy is typically measured with MPJPE and its Procrustes-aligned version, PA-MPJPE, which evaluate each frame independently. Temporal consistency is instead measured through the error of joint velocities or accelerations~\cite{kanazawa2019dynamics}, a practice that has become common in video-based human mesh recovery, where methods such as VIBE~\cite{kocabas2020vibe} and TCMR~\cite{choi2021tcmr} explicitly target temporally consistent predictions. In 3D pose estimation, temporal models improve smoothness, but frame-to-frame jitter is still observed, especially with noisy 2D inputs or frame-wise methods.

Two families of solutions have been proposed to reduce this jitter after inference. Learned refinement networks, such as SmoothNet~\cite{zeng2022smoothnet} and DeciWatch~\cite{zeng2022deciwatch}, are trained to denoise pose sequences and can be applied to the output of any estimator. However, they require training data representative of the target motions. Classical signal filters require no training and are widely used in practical systems: the One Euro Filter~\cite{casiez2012oneeuro} adapts its cutoff frequency to the speed of motion, the Savitzky--Golay filter~\cite{savitzky1964smoothing} fits local polynomials, and the median filter~\cite{tukey1977exploratory} removes isolated outliers. Despite their widespread use, these filters are typically chosen ad hoc, and their effect on the temporal consistency and spatial accuracy of different state-of-the-art architectures has not been compared.

\begin{table}[!htbp]
\centering
\footnotesize
\renewcommand{\arraystretch}{1.25}
\setlength{\tabcolsep}{4pt}
\caption{Summary of the methods and temporal filters evaluated in this work.}
\label{tab:related_work_methods}
\begin{tabularx}{\linewidth}{lll>{\raggedright\arraybackslash}Xl}
\toprule
\textbf{Method} & \textbf{Type} & \textbf{Input} & \textbf{Main idea} & \textbf{Temporal modeling} \\
\midrule
YOLO-Pose~\cite{maji2022yolopose} & 2D detector & RGB image & Single-stage pose estimation & No \\
RTMPose~\cite{jiang2024rtmpose}   & 2D detector & RGB image/crop & Top-down pose estimation & No \\
\midrule
TCPFormer~\cite{liu2025tcpformer} & Monocular 3D & 2D keypoints & Transformer with implicit pose proxy  & Yes \\
PoseMamba-L~\cite{huang2025posemamba} & Monocular 3D & 2D keypoints & Mamba-based state-space model & Yes \\
MVGFormer~\cite{liao2024mvgformer} & Multi-view 3D & Multi-view RGB + calibration & Geometry-aware Transformer  
& No \\
\midrule
One Euro Filter~\cite{casiez2012oneeuro} & Filter & 3D trajectories & Adaptive low-pass smoothing & --- \\
Savitzky--Golay~\cite{savitzky1964smoothing} & Filter & 3D trajectories & Local polynomial smoothing & --- \\
Median Filter~\cite{tukey1977exploratory} & Filter & 3D trajectories & Robust outlier removal & --- \\
\bottomrule
\end{tabularx}
\end{table}

Table~\ref{tab:related_work_methods} summarizes the methods and filters evaluated in this study, which cover 2D pose detection, monocular temporal lifting, multi-view geometric reconstruction, and temporal post-processing.

\section{Evaluation framework}
\label{se:methods}

\begin{figure}[!htbp]
\centering
\includegraphics[width=0.95\linewidth]{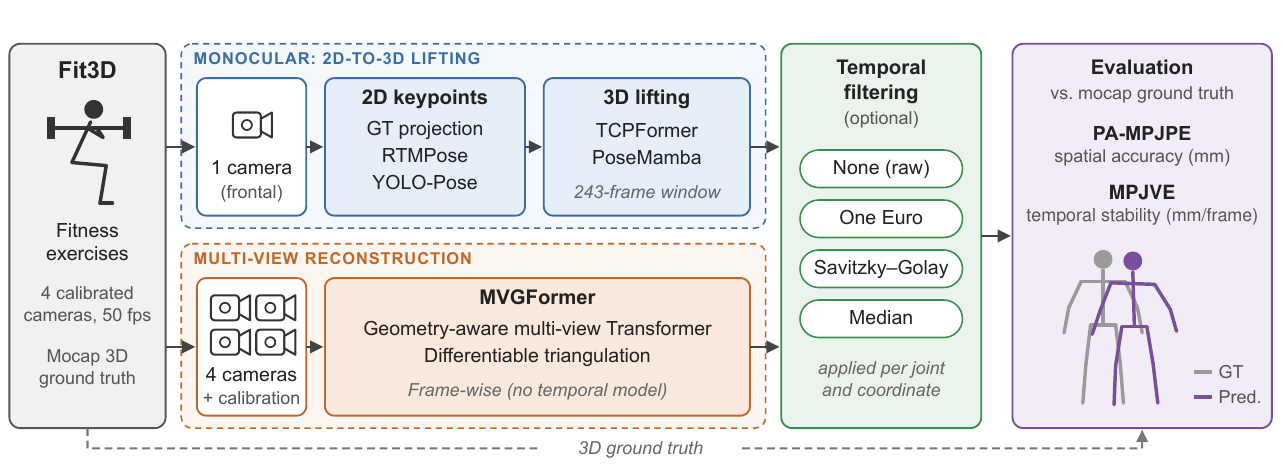}
\caption{Evaluation pipeline. Fit3D recordings feed two parallel branches: a monocular branch (top), in which 2D keypoints obtained from one frontal camera (ground-truth projection, RTMPose or YOLO-Pose) are lifted to 3D by TCPFormer or PoseMamba-L; and a multi-view branch (bottom), in which MVGFormer reconstructs the 3D pose from the four calibrated cameras. The resulting 3D trajectories are optionally smoothed with a temporal filter and evaluated against the motion-capture ground truth.}
\label{fig:pipeline}
\end{figure}

Figure~\ref{fig:pipeline} summarizes the evaluation pipeline adopted in this work. All experiments use the Fit3D dataset, with its motion-capture (Mocap) 3D ground truth as reference. For the monocular methods (in the top branch), the video from a single frontal camera is used to obtain sequences of 2D keypoints from three sources: the projection of the 3D ground truth (GT) onto the image, which serves as a reference free of detection errors, and the RTMPose and YOLO-Pose detectors. After conversion to a common joint definition, these sequences are lifted to 3D by TCPFormer and PoseMamba-L, which process temporal windows of 243 frames. 

For the multi-view method (bottom branch), images from the four cameras and their calibration parameters are processed directly by MVGFormer, which estimates each frame independently. The resulting 3D trajectories then go through the same temporal filtering stage, where they are smoothed with the One Euro, Savitzky--Golay, or Median filter, applied independently to each joint coordinate. Finally, spatial accuracy and temporal consistency are evaluated against the Mocap ground truth using PA-MPJPE and Mean Per-Joint Velocity Error (MPJVE), respectively.

All reconstructed poses, with and without temporal filtering, are compared against the Fit3D ground truth under the evaluation protocol described in Section~\ref{se:metrics}. The analysis of the 2D detector uses the original 17-joint output of TCPFormer and PoseMamba-L, whereas all comparisons involving MVGFormer, including the filtering experiments, use a common 15-joint representation (COMMON15).

\subsection{Evaluated models} 

Three representative state-of-the-art architectures are selected to cover Transformer-based temporal lifting, state-space temporal modeling, and geometry-aware multi-view reconstruction:

\begin{itemize}
    \item TCPFormer~\cite{liu2025tcpformer} is a monocular lifting architecture based on Transformer networks. The model introduces an implicit pose proxy that acts as a latent temporal representation, enabling efficient information exchange across long temporal windows. This design improves the modeling of complex motion patterns while maintaining competitive reconstruction accuracy. 
    \item PoseMamba~\cite{huang2025posemamba} replaces conventional self-attention mechanisms with selective state-space operators inspired by the Mamba architecture. This formulation substantially reduces computational complexity while preserving long-range temporal dependencies, making the model particularly attractive for long video sequences and efficient deployment. In this work, we use its large configuration, PoseMamba-L, as released by the authors. 
    \item MVGFormer~\cite{liao2024mvgformer} is the multi-view model evaluated in this work. The architecture combines deep feature extraction with differentiable triangulation, explicitly integrating geometric reasoning into the learning process. It exploits synchronized observations from multiple calibrated cameras, effectively resolving many depth ambiguities that remain challenging for monocular methods. 
\end{itemize}

Table~\ref{tab:evaluated_models} summarizes the main characteristics of the evaluated models. Both monocular methods take windows of \(T=243\) frames of 2D keypoints as input, but they differ in how temporal information is modeled. TCPFormer relies on attention, whose computational cost grows quadratically with the sequence length, whereas PoseMamba-L uses state-space scans with linear complexity. In addition, PoseMamba-L has substantially fewer parameters and requires fewer operations per frame. Since Fit3D sequences are generally longer than 243 frames, inference is performed on overlapping 243-frame windows whose predictions are then fused to reconstruct the full sequence. 

MVGFormer follows a different design: it operates directly on synchronized multi-view images and estimates each frame independently, without explicit temporal modeling. Its computational cost is also considerably higher because image features must be extracted from every camera view. All models are used with the checkpoints released by their authors, without retraining or fine-tuning on Fit3D. The computational costs reported in Table~\ref{tab:evaluated_models} correspond to the original single-window or single-frame configurations and do not include the 2D detector used by the monocular pipelines. In our overlapping-window inference, interior monocular frames are evaluated by approximately four windows and fused afterwards, so the effective inference cost per final output frame is higher than the single-window MACs reported in the table.

\begin{table}[!htbp]
\centering
\footnotesize
\renewcommand{\arraystretch}{1.25}
\setlength{\tabcolsep}{4pt}
\caption{Summary of the evaluated 3D pose estimation models. $T$ is the length of the input sequence. Parameter counts and multiply--accumulate operations (MACs) are those reported by the original authors. For TCPFormer and PoseMamba-L, MACs/frame denotes the computational cost per output frame for a 243-frame sequence. For MVGFormer, the reported MACs correspond to one frame-wise multi-view inference in the original experimental configuration.}
\label{tab:evaluated_models}

\begin{tabularx}{\linewidth}{
l
>{\raggedright\arraybackslash}X
>{\raggedright\arraybackslash}X
>{\raggedright\arraybackslash}X
r
r
>{\raggedright\arraybackslash}X
}
\toprule
\textbf{Model} &
\textbf{Backbone} &
\textbf{Input modality} &
\textbf{Temporal modeling} &
\textbf{Params} &
\textbf{MACs/frame} &
\textbf{Training} \\
\midrule
TCPFormer~\cite{liu2025tcpformer}
& Transformer with implicit pose proxy
& 2D keypoint sequence ($T=243$)
& Temporal attention, $\mathcal{O}(T^2)$
& 35.1\,M
& 449\,M
& Pretrained on Human3.6M, used as-is \\
PoseMamba-L~\cite{huang2025posemamba}
& Bidirectional spatio-temporal state-space model (Mamba)
& 2D keypoint sequence ($T=243$)
& State-space scans, $\mathcal{O}(T)$
& 6.7\,M
& 115\,M
& Pretrained on Human3.6M, used as-is \\
MVGFormer~\cite{liao2024mvgformer}
& PoseResNet-50 + Transformer decoder with triangulation
& Multi-view RGB images + camera calibration
& None (single-frame)
& 37.6\,M
& 639.7\,G
& Pretrained on CMU Panoptic, used as-is \\
\bottomrule
\end{tabularx}
\end{table}

\subsection{Input pose representation}
\label{se:input} 

\begin{figure}[!htbp]
    \centering
    \begin{subfigure}[t]{0.3117\textwidth}
        \centering
        \includegraphics[width=\linewidth]{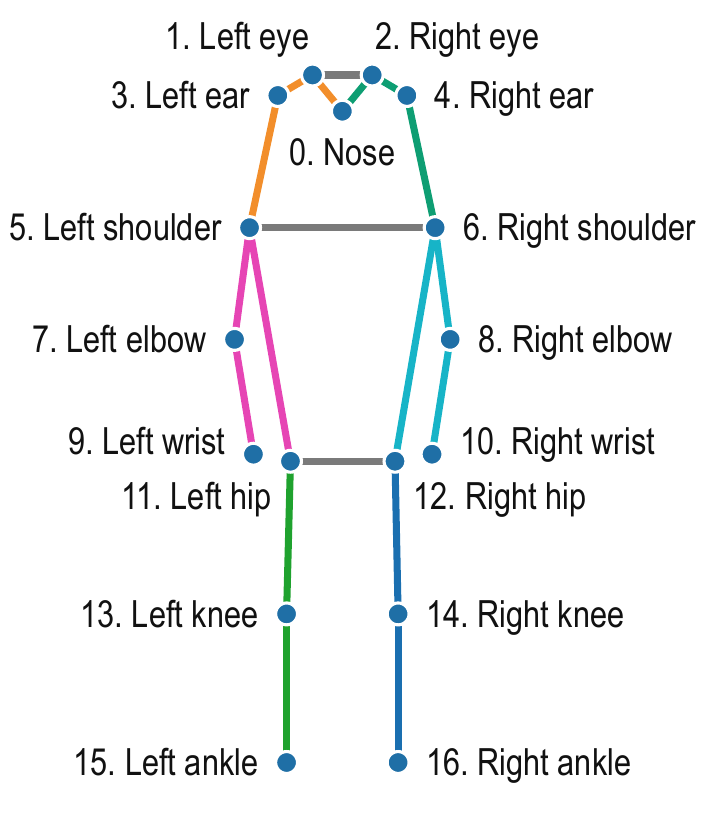}
        \caption{YOLO-Pose / RTMPose (COCO-17)}
        \label{fig:skeleton_coco17}
    \end{subfigure}
    \hfill
    \begin{subfigure}[t]{0.3291\textwidth}
        \centering
        \includegraphics[width=\linewidth]{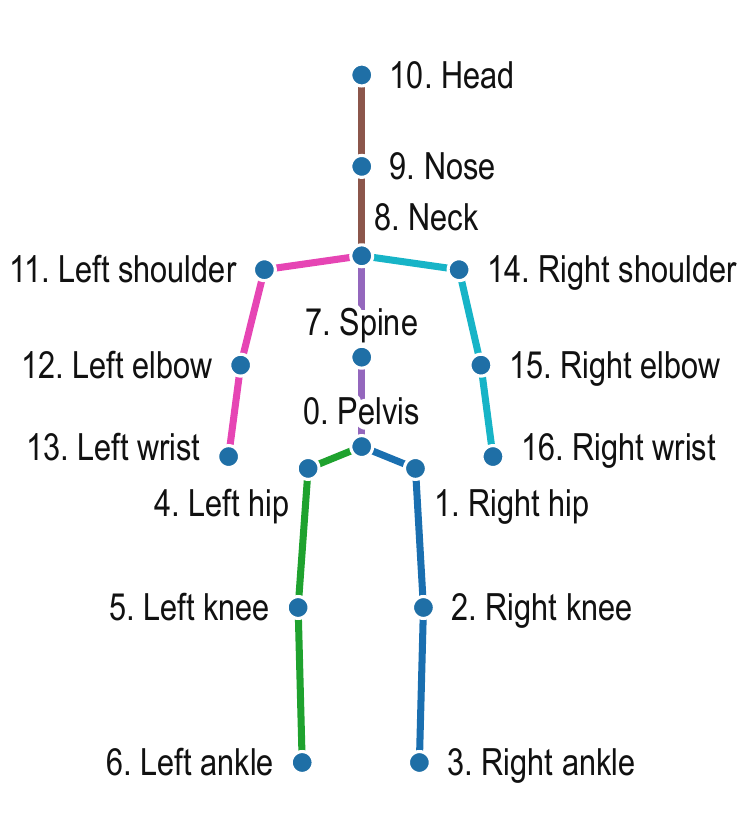}
        \caption{TCPFormer / PoseMamba-L (H36M-17)}
        \label{fig:skeleton_h36m17}
    \end{subfigure}
    \hfill
    \begin{subfigure}[t]{0.3291\textwidth}
        \centering
        \includegraphics[width=\linewidth]{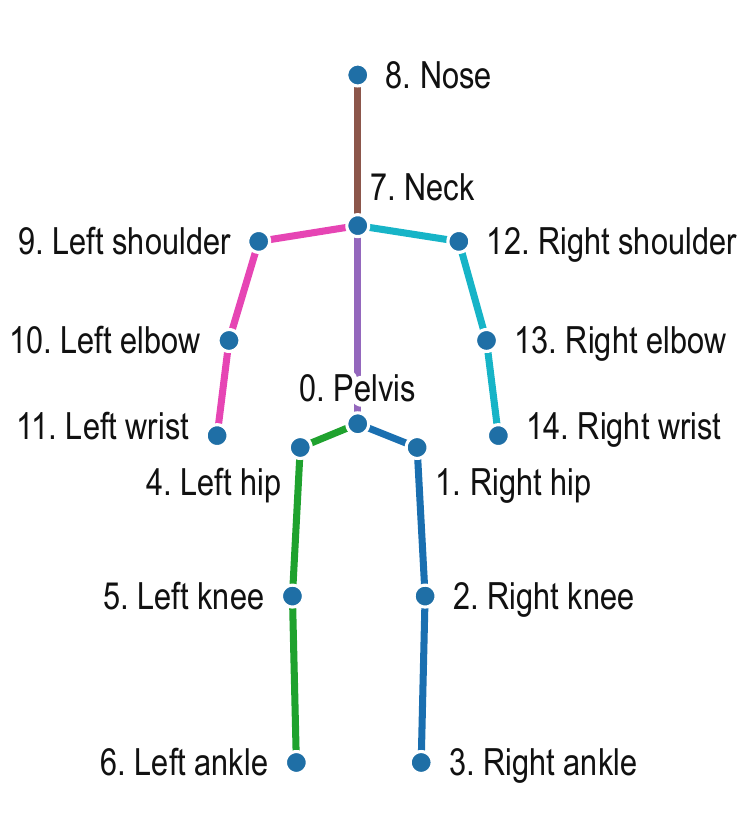}
        \caption{MVGFormer (COMMON15)}
        \label{fig:skeleton_mvg15}
    \end{subfigure}
    \caption{
     Skeleton representations used in the experimental pipeline. YOLO-Pose and RTMPose provide detections in the COCO-17 format. TCPFormer and PoseMamba-L use the 17-joint representation adopted in the monocular pipeline, which contains separate Nose and Head joints. MVGFormer follows a 15-joint Panoptic-style representation that includes Nose but no separate Head joint. For COMMON15 comparison, the monocular Spine and Head joints are therefore excluded, while Nose is retained and matched to the corresponding MVGFormer joint.
    }
    \label{fig:skeleton_representations}
\end{figure}

The monocular models receive 2D keypoints from the three sources described above: ground-truth projections, free of detection errors; RTMPose, a high-accuracy top-down detector; and YOLO-Pose, an efficient single-stage alternative suitable for real-time applications. Since the evaluated methods rely on different skeletal conventions, the input and output representations are adapted to each model, as illustrated in Fig.~\ref{fig:skeleton_representations}. 

For the monocular experiments, all 2D inputs were obtained from the same Fit3D camera (ID 65906101) to ensure an identical viewpoint for TCPFormer and PoseMamba-L. We selected this camera because it provides a predominantly frontal view of the subject, offering clear, consistent observation of the body across the evaluated exercises. Ground-truth projections, RTMPose~\cite{jiang2024rtmpose}, and YOLO-Pose~\cite{maji2022yolopose} detections were converted to the 17-joint representation expected by the lifting models. Shared joints were reordered directly, while the pelvis and neck were computed as the midpoints of the hip and shoulder pairs, respectively, and the spine as the midpoint between pelvis and neck. The resulting 2D coordinates were normalized according to the input preprocessing of the monocular models before 3D inference.

The resulting TCPFormer and PoseMamba-L predictions retain the 17-joint representation used by the monocular pipeline for their individual evaluations. For joint-wise comparison with MVGFormer, we use the common 15-joint representation containing only anatomically corresponding joints across the two skeletal conventions. MVGFormer follows the 15-joint Panoptic-style convention, which includes the neck and nose joints but does not define a separate Head joint. Therefore, it removes the spine and head joints from the 17-joint monocular representation while retaining the nose joint. In particular, the nose predicted by MVGFormer is matched to joint 9 of the monocular representation, whereas joint 10 (head) is excluded. MVGFormer predictions are then reordered from their native 15-joint ordering to COMMON15, and the Fit3D ground truth is adapted to the same convention before evaluation.

\subsection{Temporal post-processing}
\label{se:filtering}

Even models that exploit temporal context may produce small frame-to-frame fluctuations in the reconstructed trajectories, and frame-wise methods such as MVGFormer have no temporal mechanism at all. We therefore evaluate three lightweight filters that are applied after inference, independently to each coordinate $x_t$ of each joint, which represent three complementary ways of handling temporal noise:  an adaptive causal filter, a shape-preserving linear filter, and a robust non-linear filter. In the following, $\tilde{x}_t$ denotes the filtered value.

The One Euro Filter~\cite{casiez2012oneeuro} is an exponential smoother whose strength adapts to the speed of motion:
\begin{equation}
\tilde{x}_t = \alpha\,x_t + (1-\alpha)\,\tilde{x}_{t-1},
\qquad
\alpha = \frac{1}{1 + \dfrac{1}{2\pi f_c\,\Delta t}},
\qquad
f_c = f_{c,\min} + \beta\,\bigl|\widehat{\dot{x}}_t\bigr|,
\label{eq:oneeuro}
\end{equation}
where $\Delta t = 1/f_s$ and $\widehat{\dot{x}}_t$ is the velocity estimated from consecutive frames and smoothed with a low-pass filter of fixed cutoff frequency $f_d$, which prevents the noise of the derivative from causing abrupt changes in $f_c$. Slow joints get a low cutoff and strong smoothing, whereas fast joints get a higher cutoff and less lag. It is causal and inexpensive, which makes it the natural choice for real-time applications, although an isolated error may propagate to the following frames.

The Savitzky--Golay filter~\cite{savitzky1964smoothing} fits a polynomial of degree $p$ by least squares to a centered window of $N=2m+1$ frames and keeps its value at the center:
\begin{equation}
\min_{a_0,\dots,a_p}\sum_{i=-m}^{m} \Bigl( \sum_{k=0}^{p} a_k i^k-x_{t+i}\Bigr)^2,\qquad\tilde{x}_t = a_0.
\label{eq:sg_fit}
\end{equation}

This is equivalent to a convolution with fixed coefficients. Unlike a moving average ($p=0$), it preserves the local curvature, reducing jitter without flattening the peaks of the exercise movements. Being non-causal, it is best suited for offline analysis.

The Median filter~\cite{tukey1977exploratory} replaces each sample by the median of a centered window of $N=2m+1$ frames:
\begin{equation}
\tilde{x}_t = \operatorname{median}\left\{x_{t-m},\dots,x_{t+m}\right\}.
\label{eq:median}
\end{equation}

Because the median ignores extreme values, it removes isolated spikes caused by occasional misdetections or depth errors while preserving sharp transitions, although it is less effective against small continuous jitter.

We selected the parameters through a global search over predefined configurations, minimizing the mean MPJVE while limiting the increase in PA-MPJPE to 0.5\,mm relative to the original predictions. The One Euro Filter used $(f_s,f_{c,\min},\beta,f_d)=(50,10,0.01,2)$ for TCPFormer, $(50,100,0.01,1)$ for PoseMamba-L, and $(50,1,0.001,1)$ for MVGFormer. The Savitzky--Golay filter used $(N,p)=(21,4)$ for TCPFormer, $(11,4)$ for PoseMamba-L, and $(21,2)$ for MVGFormer. The Median filter used window sizes of $N=5$, $3$, and $41$ for TCPFormer, PoseMamba-L, and MVGFormer, respectively.

\section{Experiments}
\label{se:experiments}

\subsection{Dataset}
\label{se:dataset} 

The Fit3D dataset~\cite{fit3d2024} is a large-scale benchmark specifically designed for 3D human motion analysis during fitness exercises. It focuses on complex full-body movements involving large joint excursions, self-occlusions, and rapid changes of body configuration. The dataset provides synchronized RGB recordings from multiple calibrated cameras, along with accurate 3D ground-truth joint annotations obtained with a professional motion capture system.

Table~\ref{tab:fit3d-characteristics} summarizes the main characteristics, including the number of sequences and the types of exercises represented. As detailed below, our experiments use the eight subjects of the training split and 28 exercises. We excluded warm-up sequences because they are only generically described without identifying the movement performed.

The original Fit3D data collection involves 13 subjects performing 47 exercise types grouped into four families: warm-ups, barbell, dumbbell, and equipment-free exercises. In total, it comprises 611 multi-view sequences, each containing at least five repetitions of the exercise, recorded by four RGB cameras at $900\times900$ pixels and 50\,fps. The 3D ground truth was captured with a 12-camera Vicon marker-based system and is provided as 25-joint skeletons that include the 17 joints of the Human3.6M convention. 

\begin{table}[!htbp]
\centering
\footnotesize
\renewcommand{\arraystretch}{1.25}
\setlength{\tabcolsep}{4pt}
\caption{Main characteristics of the Fit3D dataset.}
\label{tab:fit3d-characteristics}
\begin{tabularx}{\linewidth}{lX}
\toprule
\textbf{Characteristic} & \textbf{Fit3D}~\cite{fit3d2024} \\
\midrule
Application domain & Fitness training. Multiple subjects exercising. \\
Environment  & Indoor laboratory \\
Subjects & 13 subjects in the original data collection\\
Activities & 47 exercise types, in 4 families (Warmups, Barbell Exercises, Dumbbell Exercises, Equipment-Free Exercises) \\
Sequences & 611 multi-view sequences ($\geq 5$ repetitions each), i.e., 2444 videos across the four cameras. \\ 
RGB cameras & 4 (multiple cameras) \\
Image resolution & $900 \times 900$ px \\
Video frame rate & 50\,fps \\
MoCap system & Vicon, 12 cameras. \\
3D skeleton & Ground-truth 3D skeletons with 25 joints (includes the 17 Human3.6M joints) \\
Camera calibration & Intrinsics and extrinsics \\
Official split & Train: 8 subjects; Test: 3 subjects \\
Main challenges & Large range of motion, non-upright poses, equipment occlusions \\
\bottomrule
\end{tabularx}
\end{table}

The publicly released Fit3D benchmark defines an official split comprising 8 training subjects and 3 test subjects. However, the 3D ground truth of the test split is withheld for the online benchmark. For this reason, all experiments in this work use the sequences of the training split (subjects s03, s04, s05, s07, s08, s09, s10, and s11), which provide the four camera views together with the complete 3D annotations. 

\begin{figure}[!htbp]
    \centering
    \begin{tabular}{cccc}
        \includegraphics[width=0.22\textwidth]{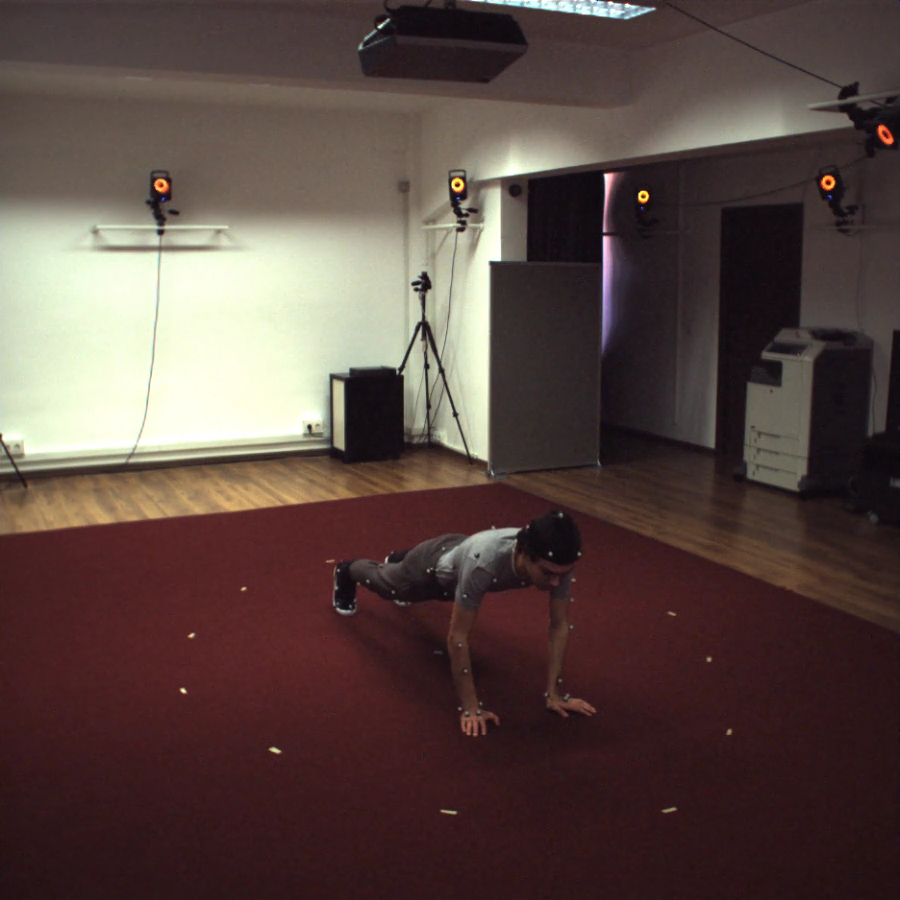} & 
        \includegraphics[width=0.22\textwidth]{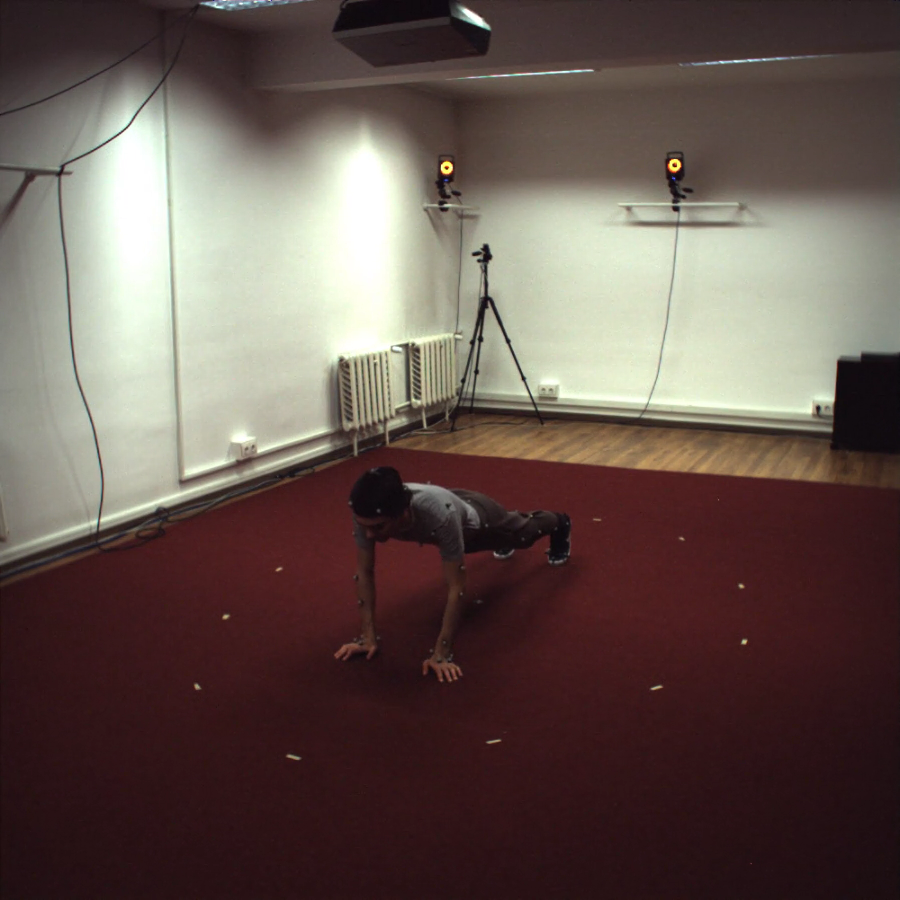} & 
        \includegraphics[width=0.22\textwidth]{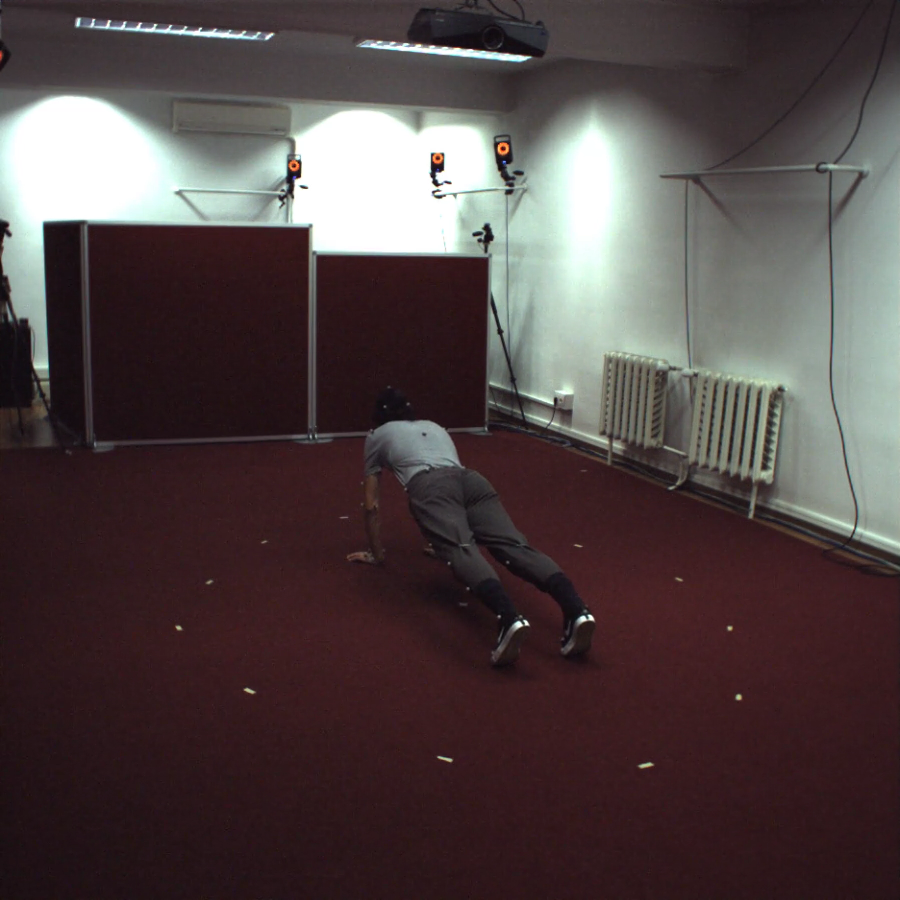} & 
        \includegraphics[width=0.22\textwidth]{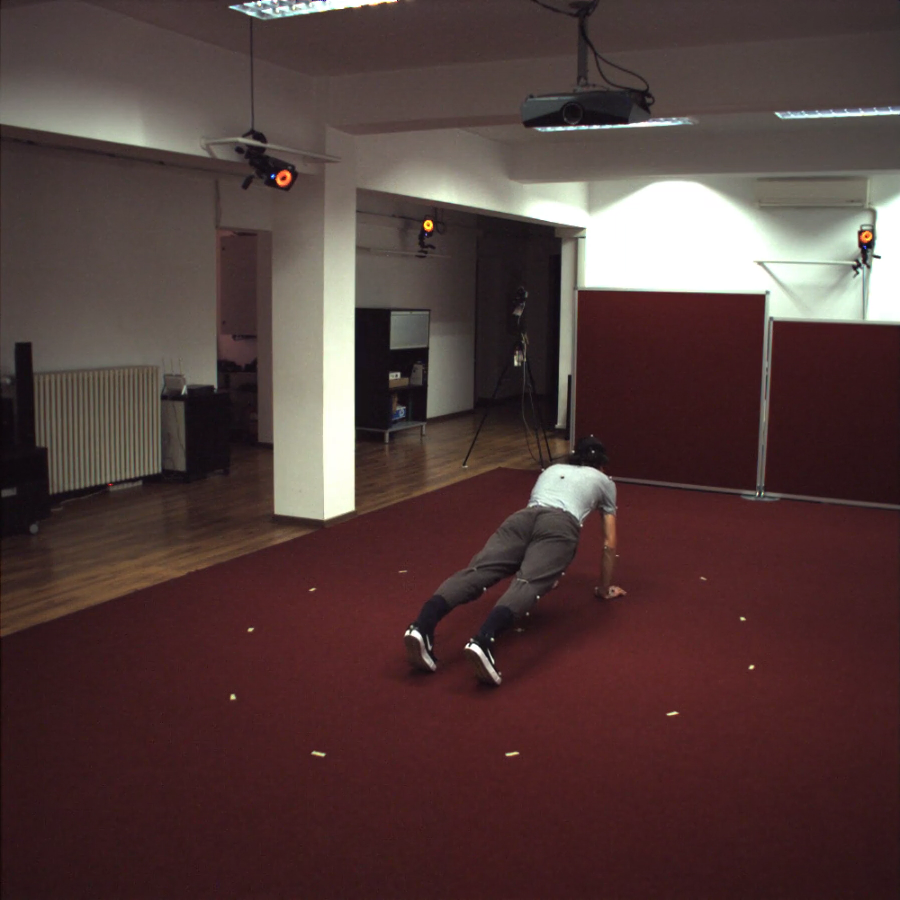} \\
        \includegraphics[width=0.22\textwidth]{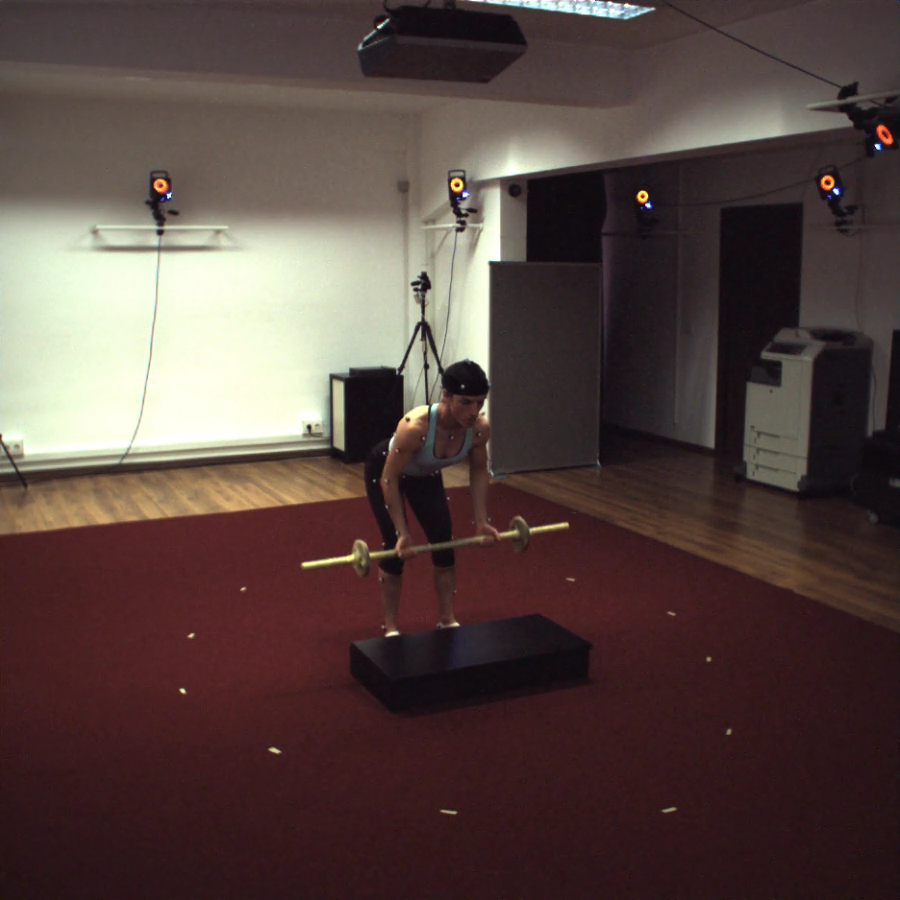} &  
        \includegraphics[width=0.22\textwidth]{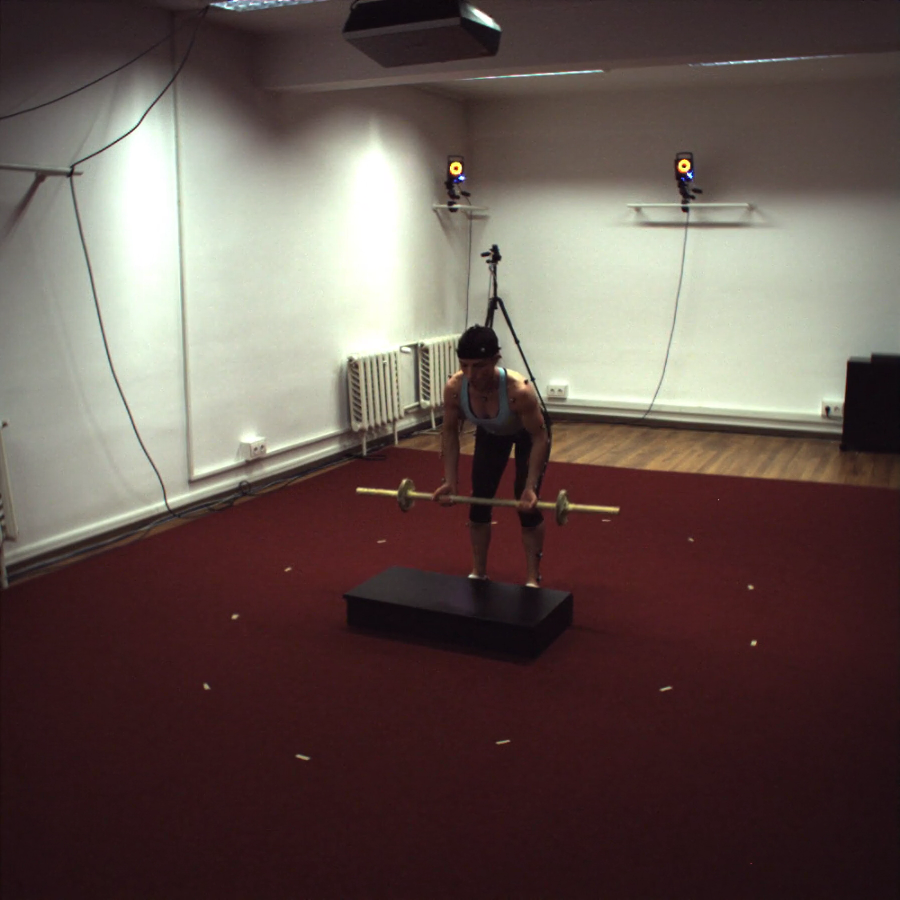} &
        \includegraphics[width=0.22\textwidth]{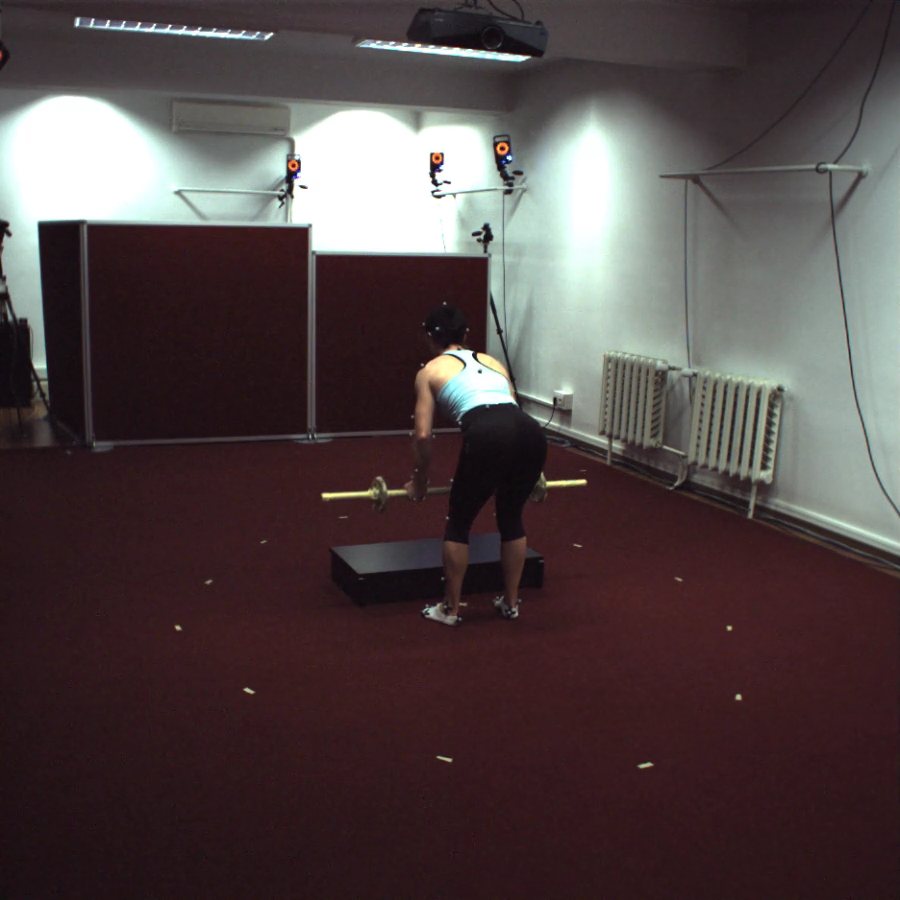} &  
        \includegraphics[width=0.22\textwidth]{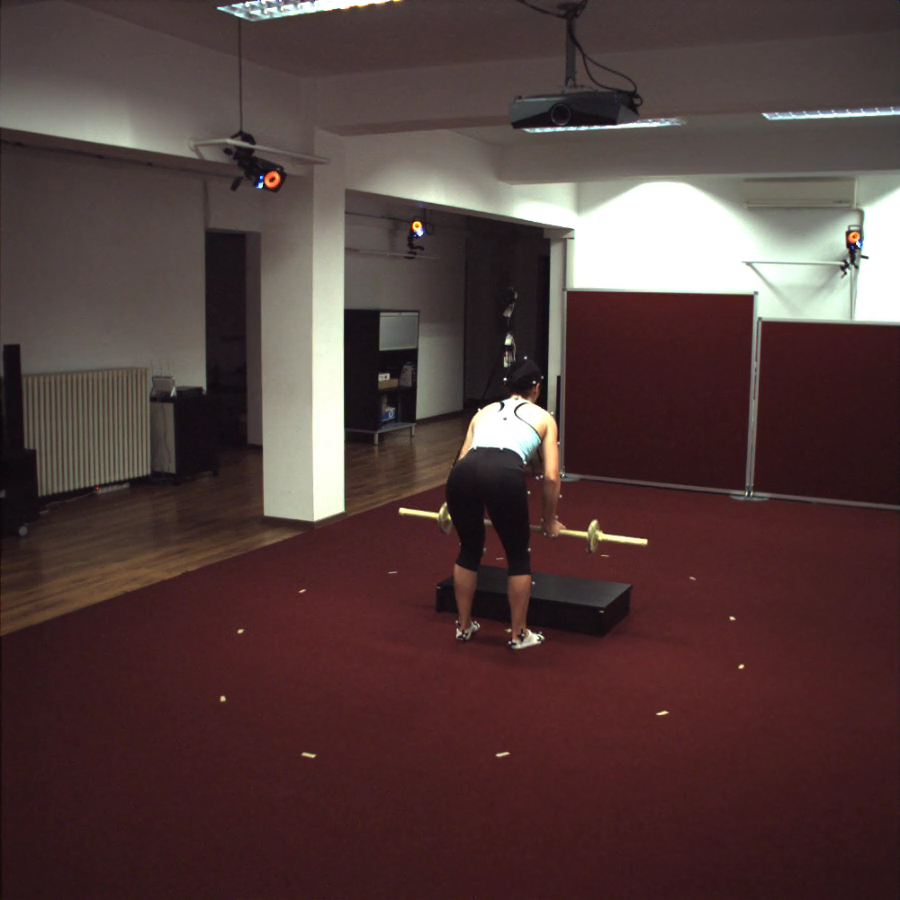}\\        
    \end{tabular}
    \caption{Fit3D dataset: Images of the four cameras of frame 400 from the s04 and s08 sequences of the pushup and drag curl exercises, respectively.}
    \label{fig:fit3d_exercises}
\end{figure}

Figure~\ref{fig:fit3d_exercises} shows images of frame 400 from the four cameras, corresponding to the s04 and s08 sequences of the pushup and drag curl exercises, respectively. These examples illustrate the multi-view observations available for each exercise. In addition to RGB images and 3D annotations, the dataset provides camera calibration parameters, which we used to project the 3D joints onto one of the frontal cameras for the monocular methods and to feed the MVGFormer method directly.

\subsection{Evaluation metrics}
\label{se:metrics}

We evaluate two complementary aspects of the predictions: the spatial accuracy of each pose and the temporal consistency of the sequence.

We use the Procrustes-Aligned Mean Per Joint Position Error (PA-MPJPE) to measure spatial accuracy. We do not use plain MPJPE because none of the models were trained on Fit3D, so their outputs may differ from the ground truth in scale, orientation, or reference frame. For each frame $t$, the prediction $\hat{X}_t$ is first aligned to the ground truth $X_t$ with the similarity transform $A_t(\hat{X}_t) = s_t R_t \hat{X}_t + \mathbf{b}_t$ obtained by Procrustes analysis, and the error is averaged over $T$ frames and $J$ joints:
\begin{equation}
\mathrm{PA\text{-}MPJPE} = \frac{1}{TJ}\sum_{t=1}^{T}\sum_{j=1}^{J}
\big\lVert A_t(\hat{X}_{t,j}) - X_{t,j} \big\rVert_2 .
\label{eq:pampjpe}
\end{equation}

To analyze the influence of the 2D detector, we compute PA-MPJPE on the original 17-joint representation used by TCPFormer and PoseMamba-L. In contrast, we compare monocular and multi-view models and evaluate temporal filtering using COMMON15. In these experiments, the predictions and ground truth are converted to the same 15-joint representation before applying Procrustes alignment. We also report PA-MPJPE per joint, averaging over time, to identify which body parts concentrate the reconstruction error.

Since PA-MPJPE evaluates each frame independently, it does not capture temporal inconsistencies such as jitter or abrupt joint displacements. Temporal consistency is therefore evaluated with a separate protocol, applied identically to all models and filtering configurations. At every frame, both prediction and ground truth are expressed in millimeters and centered relative to the pelvis,
\begin{equation}
Y_{t,j}=X_{t,j}-X_{t,\mathrm{pelvis}}, \qquad
\hat{Y}_{t,j}=\hat{X}_{t,j}-\hat{X}_{t,\mathrm{pelvis}}.
\label{eq:preproc}
\end{equation}
Since the pelvis is subtracted at each frame, the translation component of the camera extrinsics cancels out. Consequently, the metric describes root-relative articulated motion rather than the global displacement of the subject through the scene. The pelvis has zero error by construction and is included in the average over the common 15 joints; since this affects all models and filters by the same factor, it does not change their relative comparison.

Unlike PA-MPJPE, the temporal evaluation does not apply frame-wise Procrustes alignment or estimate an additional body-size scale factor. The predictions are evaluated in their corresponding calibrated reference frames, preserving their original temporal displacements. Applying an independent similarity transformation at each frame could introduce artificial differences between consecutive poses, thereby affecting the velocity error.

Temporal consistency is measured using the Mean Per-Joint Velocity Error (MPJVE). For each joint $j$, the frame-to-frame displacement is defined as $\Delta Y_{t,j}=Y_{t,j}-Y_{t-1,j}$ (and analogously $\Delta\hat{Y}_{t,j}$), and
\begin{equation}
\mathrm{MPJVE}=
\frac{1}{(T-1)J}
\sum_{t=2}^{T}
\sum_{j=1}^{J}
\left\|
\Delta \hat{Y}_{t,j}-\Delta Y_{t,j}
\right\|_2 .
\label{eq:mpjve}
\end{equation}

The metric is reported in mm/frame and captures differences in both motion magnitude and direction. Results are first averaged over all frames and joints of each sequence and then aggregated across sequences and subjects using the same protocol for all models and filtering configurations.

For visualization only, we additionally compute the mean body velocity,
\begin{equation}
\bar{v}_t(Y)=
\frac{1}{J}
\sum_{j=1}^{J}
\left\|
\Delta Y_{t,j}
\right\|_2,
\label{eq:velocity}
\end{equation}
which is computed analogously for the predicted and filtered sequences. For these plots, a single body-size scale factor is estimated per sequence from the ratio between the mean bone lengths of the ground truth and the original prediction, and applied to the original and all filtered predictions. This normalization is not used when computing MPJVE.

\subsection{Implementation details}
\label{sec:implementation}

The experiments were developed in Google Colab. The experimental environment used Python v3.12.13 and PyTorch v2.10.0 with CUDA v12.8 support, running mainly on a Tesla T4 GPU. PyTorch was used for deep learning inference, while NumPy, OpenCV, Pandas, Matplotlib, and SciPy were used for pose processing, video and image handling, result aggregation, visualization, and temporal filtering.

For TCPFormer and PoseMamba-L, inference over complete Fit3D sequences is performed using sliding temporal windows of 243 frames with a stride of 60 frames, corresponding to an overlap of 183 frames between consecutive windows. Each interior frame is therefore predicted by approximately four overlapping windows. The predictions contributed by different windows are combined using a strictly positive Hann-based weighting function based on the Hann window~\cite{harris1978windows}, which assigns greater weight to the central region of each window while reducing the influence of its boundaries. For each frame, the weighted predictions are accumulated and normalized by the sum of their corresponding weights.

The final window is explicitly anchored at the end of the sequence to ensure that all frames are covered. When a sequence is shorter than 243 frames, it is padded by repeating its final frame until the required window length is reached; the padded predictions are discarded after inference. Root-centering with respect to the pelvis is applied after the overlapping predictions have been fused. This strategy is intended to reduce boundary discontinuities between independently processed temporal windows and to produce a single continuous 3D trajectory for each sequence.

For MVGFormer, the native \(900\times900\) Fit3D frames were transformed to the \(960\times540\) network input size using an aspect-ratio-preserving affine transformation. For the square Fit3D images, this operation scales the image content uniformly and centers it within the \(960\times540\) input canvas, introducing lateral padding rather than anisotropic resizing. The original intrinsic and extrinsic camera parameters were retained, while the forward and inverse affine transformations were provided to the model as part of the view metadata. 

The code used to preprocess the Fit3D dataset, run inference with the three evaluated models, and reproduce all reported metrics and figures, together with some of the 3D predictions used in the temporal smoothing experiments, is publicly available at \url{https://github.com/Nestpr/fit3d-3d-pose-benchmark} under the Apache~2.0 License.

\subsection{Influence of the 2D detector}
\label{se:results}

This experiment evaluates TCPFormer and PoseMamba-L with the three 2D input sources: ground-truth 2D keypoints, YOLO-Pose, and RTMPose detections.

\begin{figure}[!htbp]
    \centering
    \includegraphics[width=0.85\textwidth]{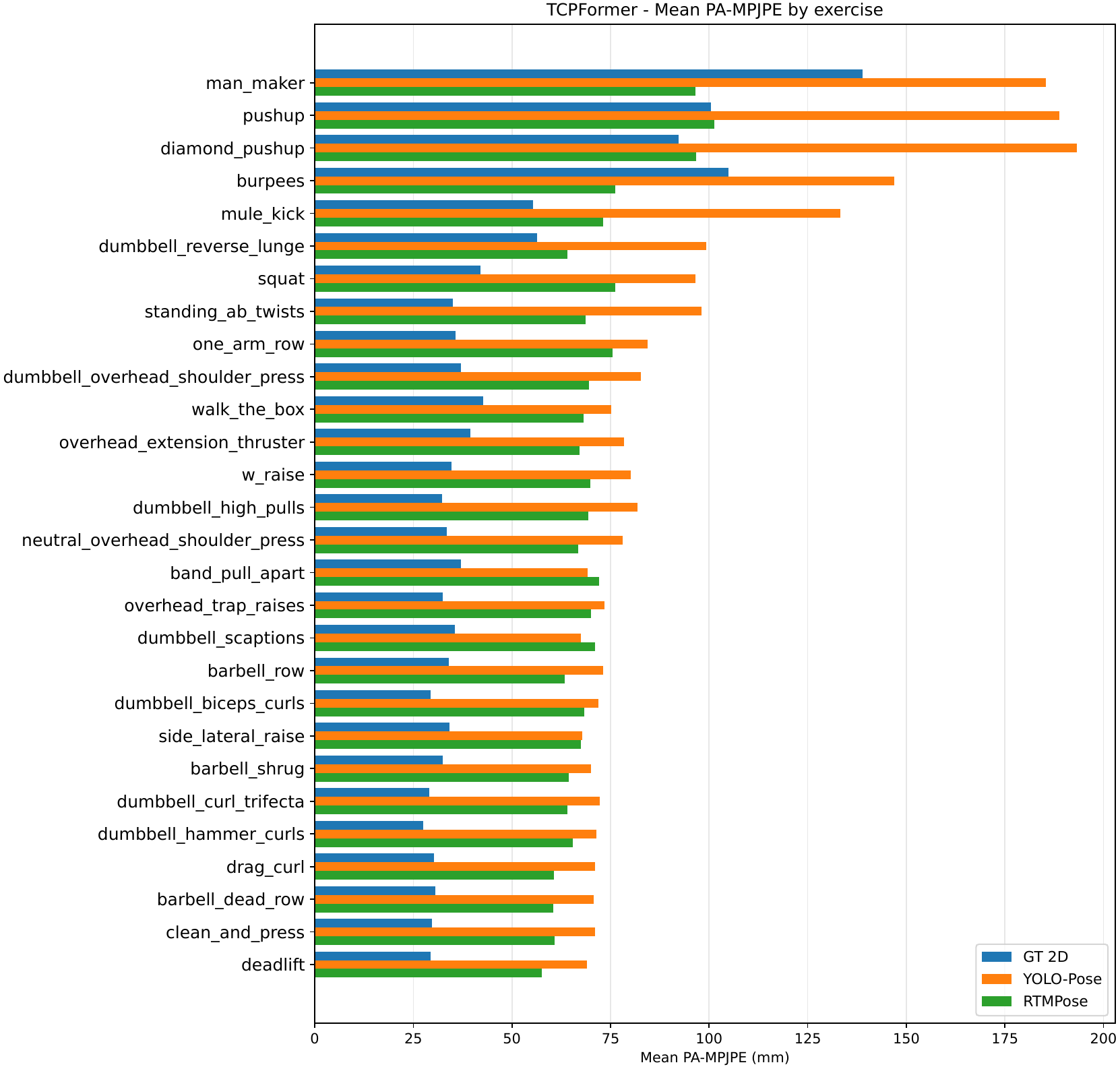}
    \caption{Mean PA-MPJPE per exercise for TCPFormer according to the 2D keypoint source.}
    \label{fig:tcpformer_detector_exercises}
\end{figure}

\begin{figure}[!htbp]
    \centering
    \includegraphics[width=0.85\textwidth]{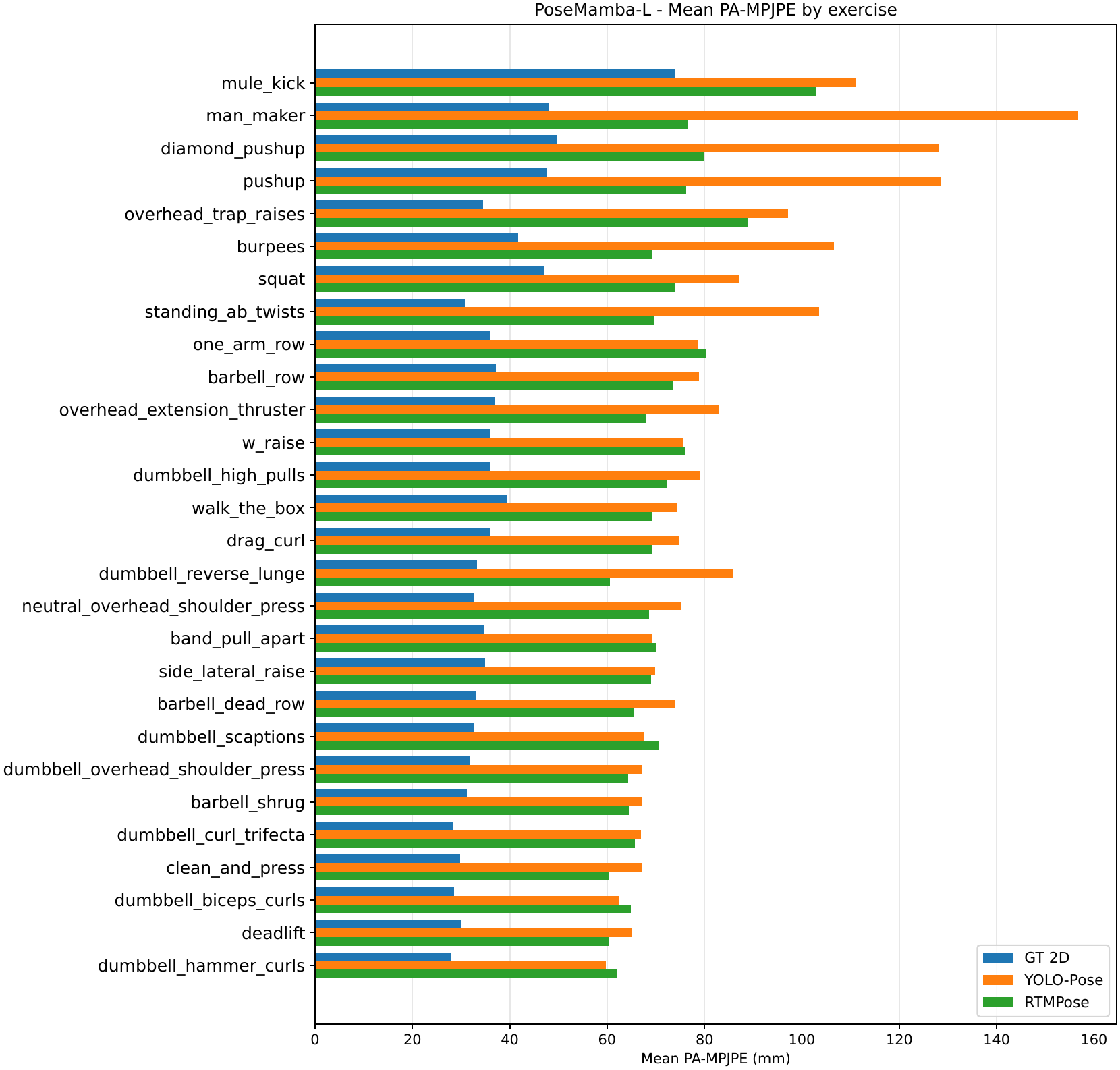}
    \caption{Mean PA-MPJPE per exercise for PoseMamba-L according to the 2D keypoint source.}
    \label{fig:posemamba_detector_exercises}
\end{figure}

Figs.~\ref{fig:tcpformer_detector_exercises} and~\ref{fig:posemamba_detector_exercises} show the PA-MPJPE obtained by TCPFormer and PoseMamba-L for each exercise under the three input configurations. On average, ground-truth 2D keypoints yield the lowest PA-MPJPE for both monocular models. However, this trend is not uniform across all exercises, particularly for TCPFormer in some challenging floor-supported actions, where RTMPose can achieve comparable or lower errors.

When automatic 2D detections are used, the reconstruction error increases substantially. YOLO-Pose produces the highest errors for most exercises in both models, whereas RTMPose reduces the error in most cases, confirming that errors in the 2D stage propagate directly to the lifted 3D pose.

The sensitivity to the 2D detector is particularly visible in exercises involving complex body configurations, ground contact, or rapid changes in posture, such as \textit{man maker}, \textit{pushup}, \textit{diamond pushup}, \textit{burpees}, and \textit{mule kick}. These movements are more challenging for 2D detectors due to self-occlusions, non-upright poses, and large variations in limb visibility. In contrast, exercises with more upright or regular motion patterns, such as \textit{deadlift}, \textit{barbell dead row}, \textit{clean and press}, or \textit{dumbbell curl} variants, generally lead to lower PA-MPJPE values.

\begin{table}[!htbp]
\centering
\caption{Mean PA-MPJPE (mm) per subject for TCPFormer and PoseMamba-L using three 2D input sources, evaluated on their original 17-joint representation.}
\label{tab:pampjpe_monocular}
\resizebox{\linewidth}{!}{%
\begin{tabular}{lccccccccc}
\toprule
Model / input & s03 & s04 & s05 & s07 & s08 & s09 & s10 & s11 & \textbf{Mean} \\
\midrule
TCPFormer -- GT 2D
& 42.17 & 50.02 & 42.02 & 45.91 & 52.15 & 39.92 & 49.56 & 47.50
& \textbf{46.16} \\
TCPFormer -- YOLO-Pose
& 87.52 & 94.19 & 95.21 & 103.65 & 95.49 & 90.16 & 97.91 & 84.71
& \textbf{93.61} \\
TCPFormer -- RTMPose
& 69.57 & 74.63 & 71.88 & 70.94 & 71.23 & 71.96 & 69.70 & 67.20
& \textbf{70.89} \\
\addlinespace
PoseMamba-L -- GT 2D
& 34.87 & 45.79 & 34.07 & 36.76 & 38.78 & 31.30 & 39.83 & 35.57
& \textbf{37.12} \\
PoseMamba-L -- YOLO-Pose
& 78.68 & 85.83 & 82.51 & 94.17 & 85.16 & 80.35 & 92.54 & 75.32
& \textbf{84.32} \\
PoseMamba-L -- RTMPose
& 71.55 & 77.07 & 69.36 & 69.05 & 71.75 & 72.13 & 69.43 & 68.81
& \textbf{71.14} \\
\bottomrule
\end{tabular}}
\end{table}

Table~\ref{tab:pampjpe_monocular} summarizes the mean PA-MPJPE per subject for both monocular models using their original 17-joint representation. With ground-truth 2D keypoints, PoseMamba-L achieves a lower mean error than TCPFormer (37.12\,mm versus 46.16\,mm). A similar trend is observed with YOLO-Pose, where PoseMamba-L obtains 84.32\,mm compared with 93.61\,mm for TCPFormer. However, when RTMPose detections are used, TCPFormer achieves a slightly lower mean error (70.89\,mm versus 71.14\,mm). These results indicate that the relative performance of the two lifting architectures depends not only on the overall quality of the 2D input, but also on how each model handles the detection errors and temporal characteristics introduced by the detector.

\subsection{Monocular vs. multi-view comparison}

We next compare the monocular and multi-view approaches using the common 15-joint representation. For TCPFormer and PoseMamba-L, RTMPose is used as the 2D input, while MVGFormer performs reconstruction directly from synchronized images and camera calibration parameters.

Table~\ref{tab:pampjpe} summarizes the mean PA-MPJPE obtained for each subject and model under this common evaluation protocol. MVGFormer achieves the lowest average error, with a mean PA-MPJPE of 66.691\,mm, followed by PoseMamba-L with 68.768\,mm and TCPFormer with 69.566\,mm. The difference between the two monocular models is only 0.797\,mm, indicating comparable spatial accuracy when using RTMPose detections. MVGFormer improves upon PoseMamba-L by 2.078\,mm and TCPFormer by 2.875\,mm. Although these differences are relatively small and MVGFormer does not obtain the lowest error for every subject, the multi-view model achieves the lowest overall reconstruction error, consistent with the benefits of combining multiple calibrated observations to reduce monocular depth ambiguity.

\begin{table}[!htbp]
\centering
\caption{Mean PA-MPJPE (mm) per subject for TCPFormer, PoseMamba-L, and MVGFormer using the common 15-joint representation. The monocular models use RTMPose as 2D input.}
\label{tab:pampjpe}
\resizebox{\linewidth}{!}{%
\begin{tabular}{lccccccccc}
\toprule
Model / input & s03 & s04 & s05 & s07 & s08 & s09 & s10 & s11 & \textbf{Mean} \\
\midrule
TCPFormer -- RTMPose
& 68.394 & 71.662 & 71.473 & 70.798 & 72.220 & 70.441 & 67.010 & 64.527
& \textbf{69.566} \\

PoseMamba-L -- RTMPose
& 68.922 & 74.660 & 68.557 & 68.321 & 71.521 & 69.209 & 64.481 & 64.476
& \textbf{68.768} \\

MVGFormer (multi-view)
& 61.672 & 60.299 & 67.718 & 70.701 & 77.136 & 71.240 & 55.807 & 68.952
& \textbf{66.691} \\
\bottomrule
\end{tabular}}
\end{table}

\begin{figure}[!htbp]
    \centering
    \includegraphics[width=1.0\textwidth]{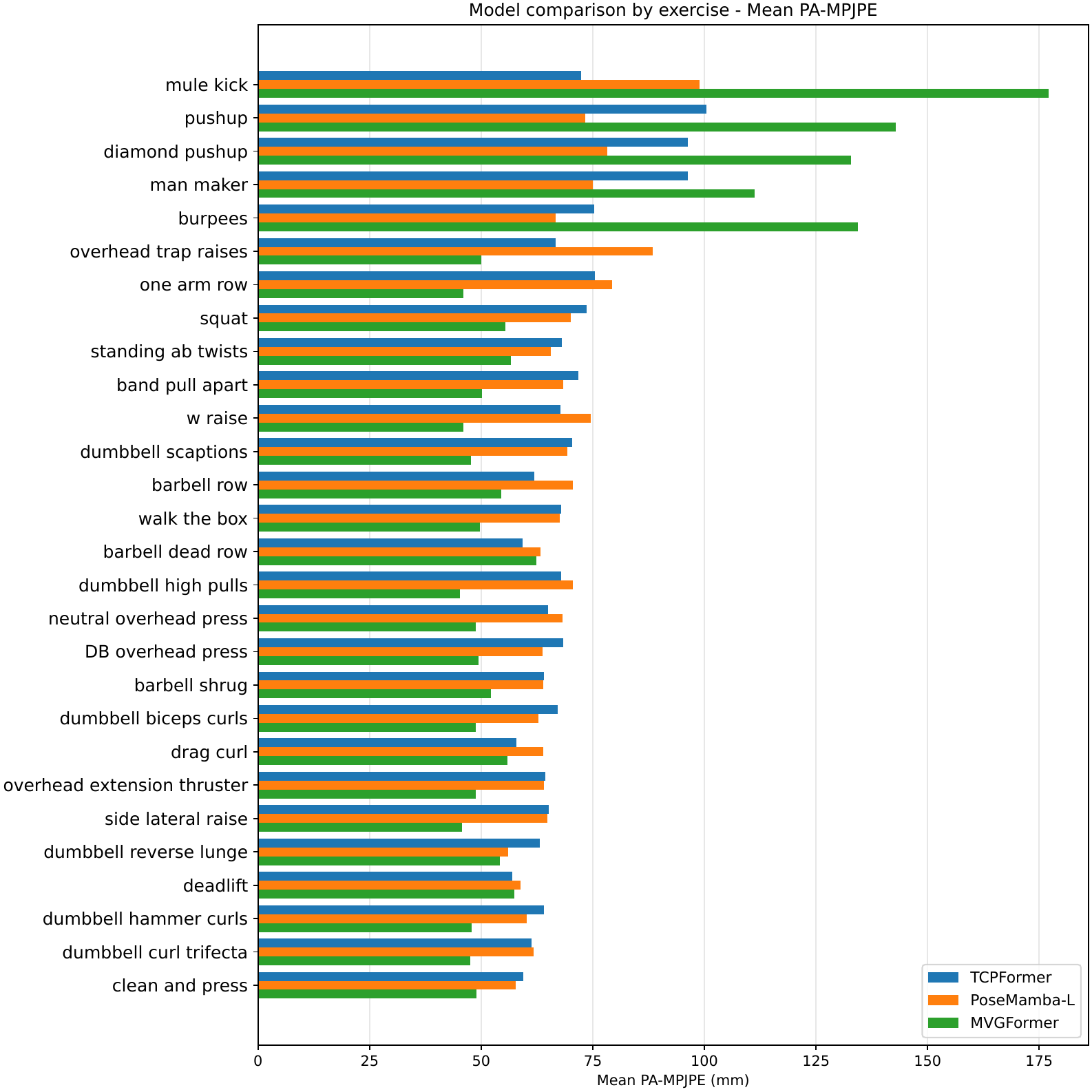}
    \caption{Comparison of the mean PA-MPJPE per exercise for TCPFormer, PoseMamba-L, and MVGFormer. TCPFormer and PoseMamba-L use RTMPose as 2D input.}
    \label{fig:comparison_exercise_error}
\end{figure}

\begin{figure}[!htbp]
    \centering
    \renewcommand{\arraystretch}{1.2}
    \setlength{\tabcolsep}{3pt}
    \begin{tabular}{
        >{\centering\arraybackslash}m{0.08\linewidth}
        >{\centering\arraybackslash}m{0.21\linewidth}
        >{\centering\arraybackslash}m{0.21\linewidth}
        >{\centering\arraybackslash}m{0.21\linewidth}
        >{\centering\arraybackslash}m{0.21\linewidth}
    }
        \toprule
        \textbf{Frame} & \textbf{Ground truth} & \textbf{TCPFormer} & \textbf{PoseMamba-L} & \textbf{MVGFormer} \\
        \midrule
        \textbf{568} &
        \makebox[\linewidth][c]{\includegraphics[width=0.95\linewidth]{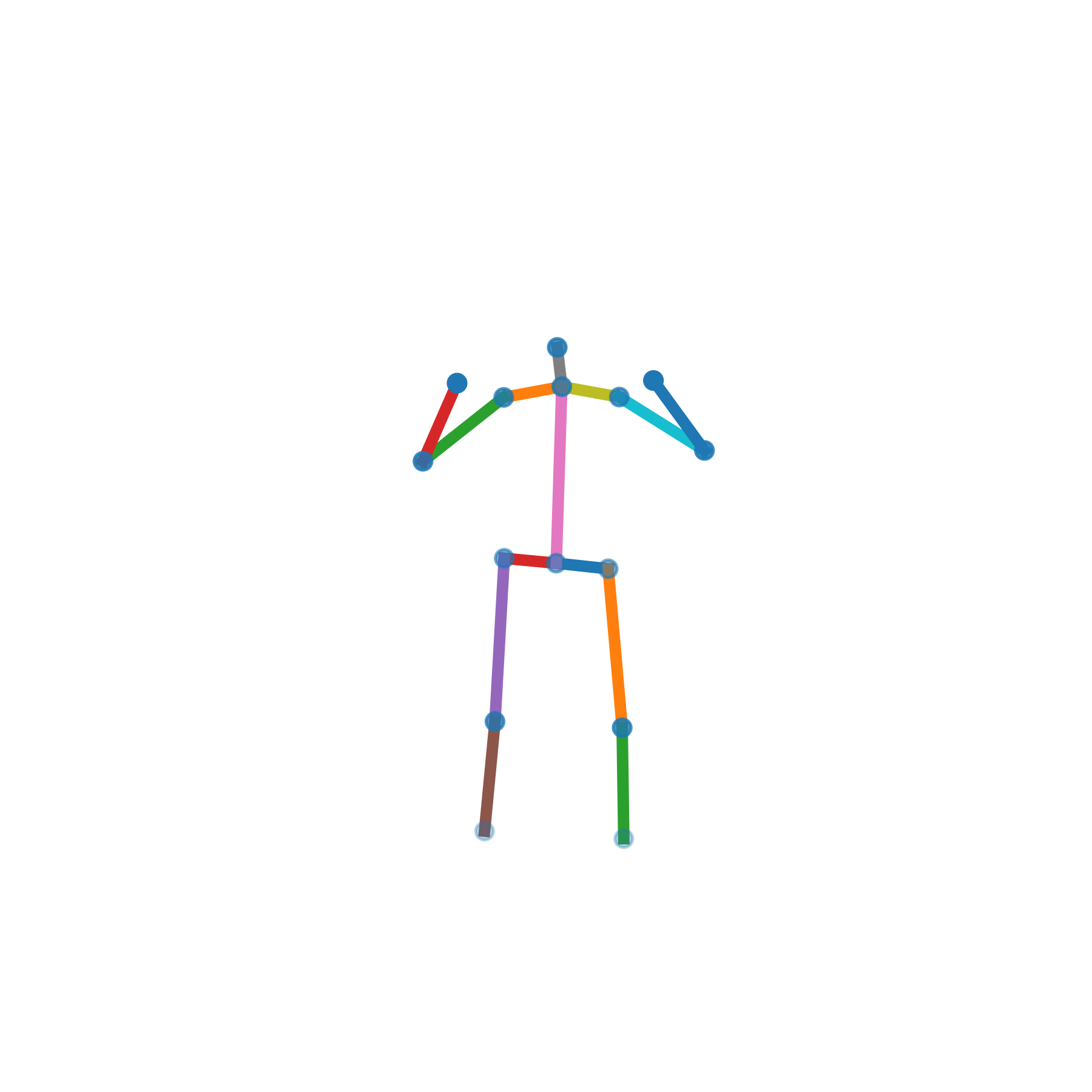}} &
        \makebox[\linewidth][c]{\includegraphics[width=0.95\linewidth]{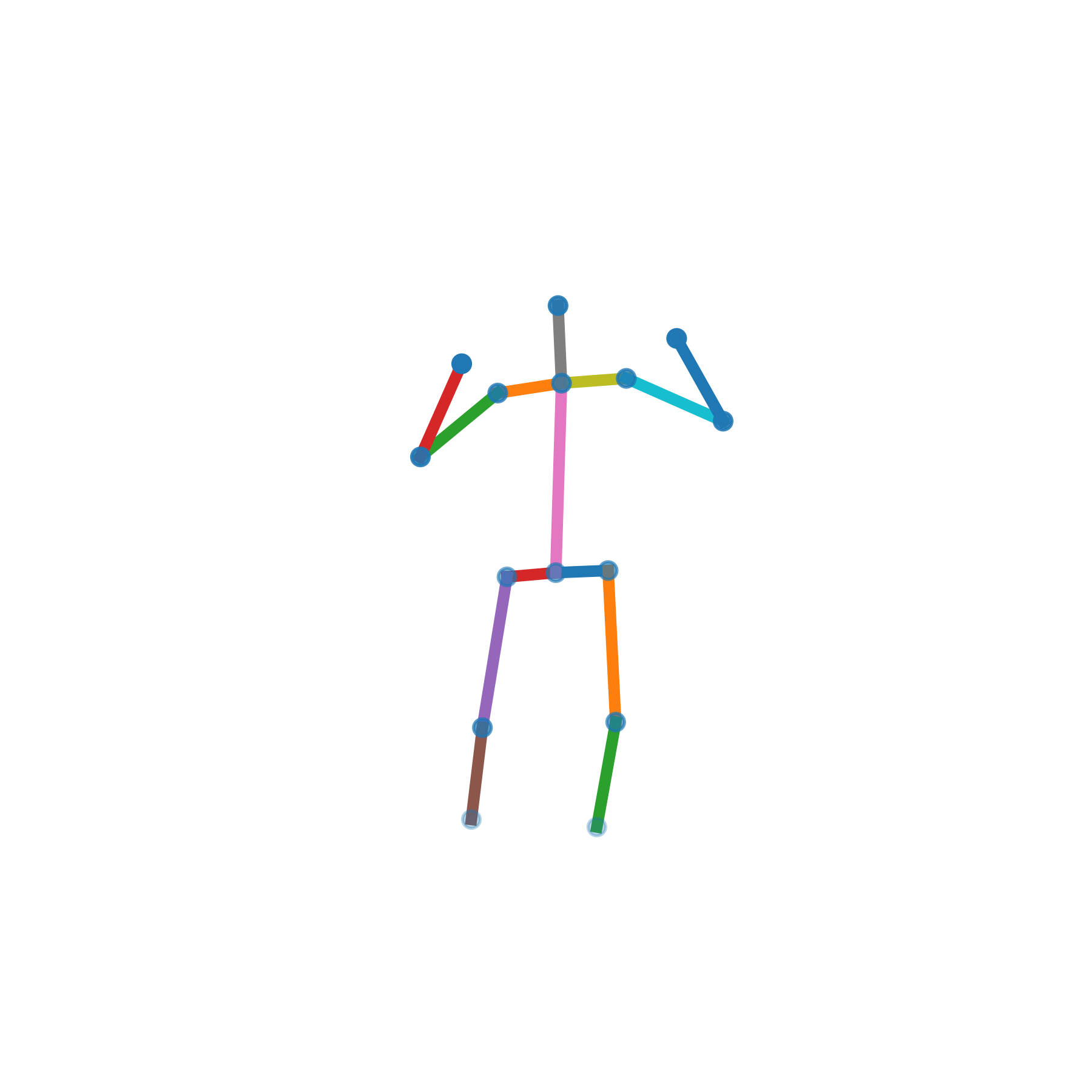}} &
        \makebox[\linewidth][c]{\includegraphics[width=0.95\linewidth]{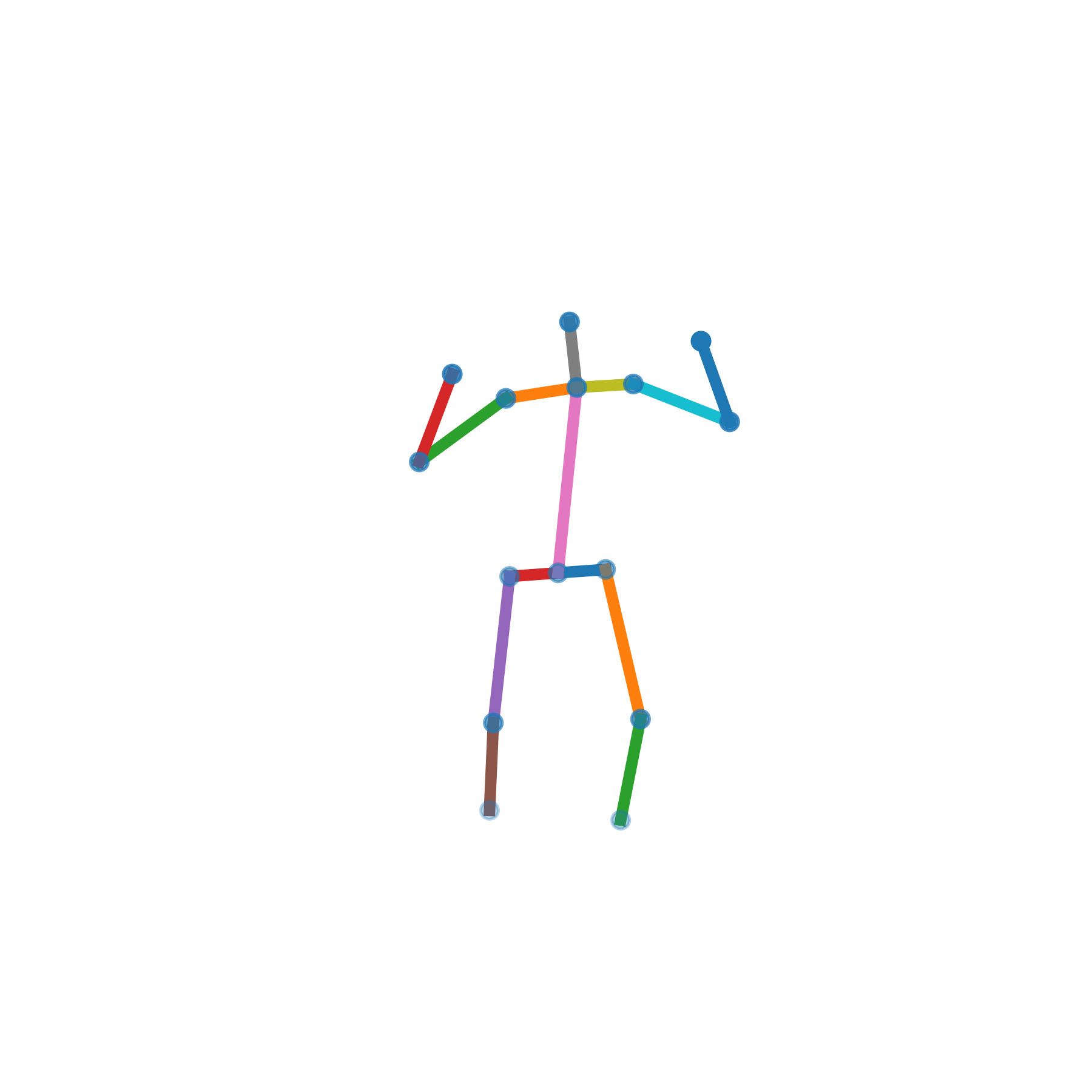}} &
        \makebox[\linewidth][c]{\includegraphics[width=0.95\linewidth]{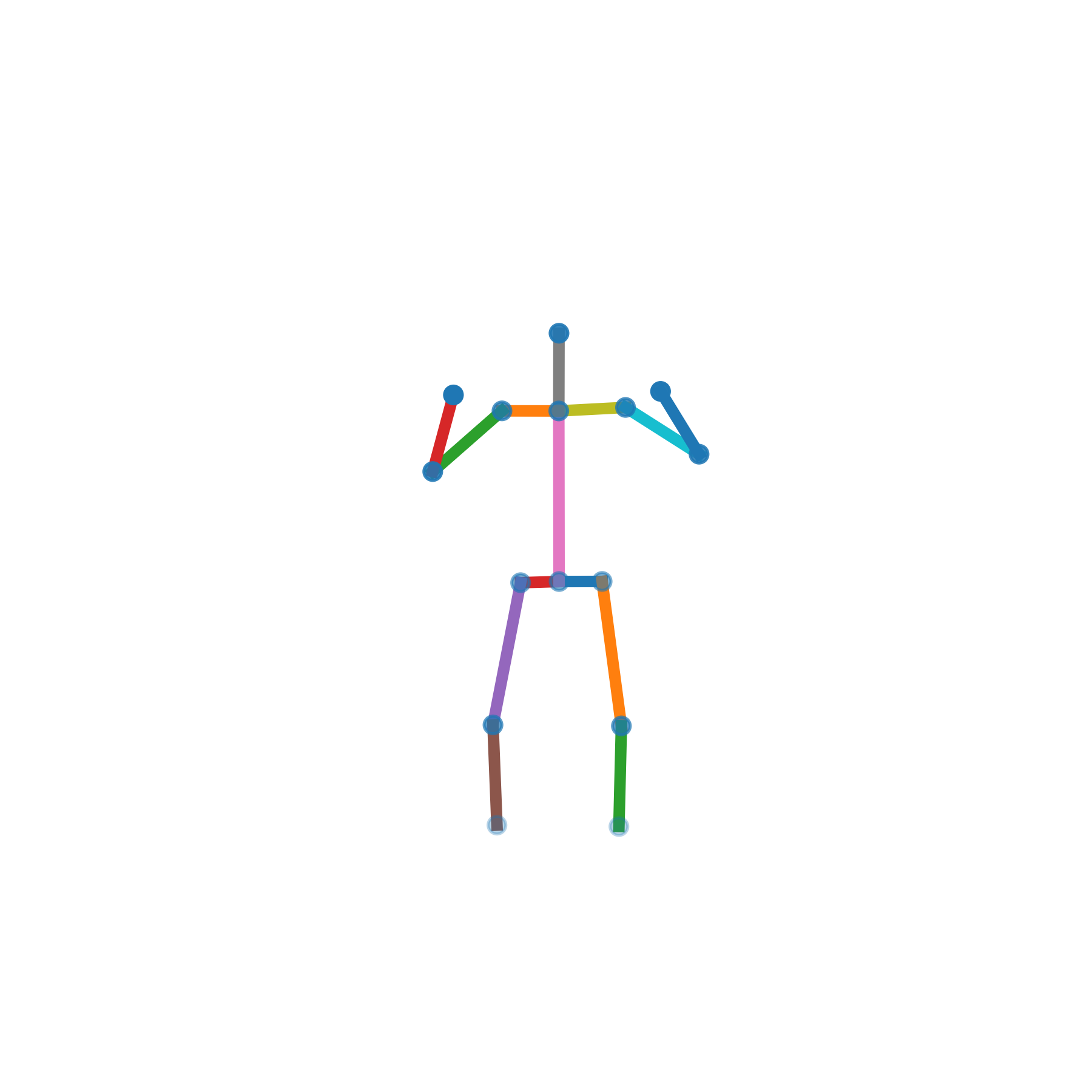}} \\[0.15cm]
        \textbf{1854} &
        \makebox[\linewidth][c]{\includegraphics[width=0.82\linewidth]{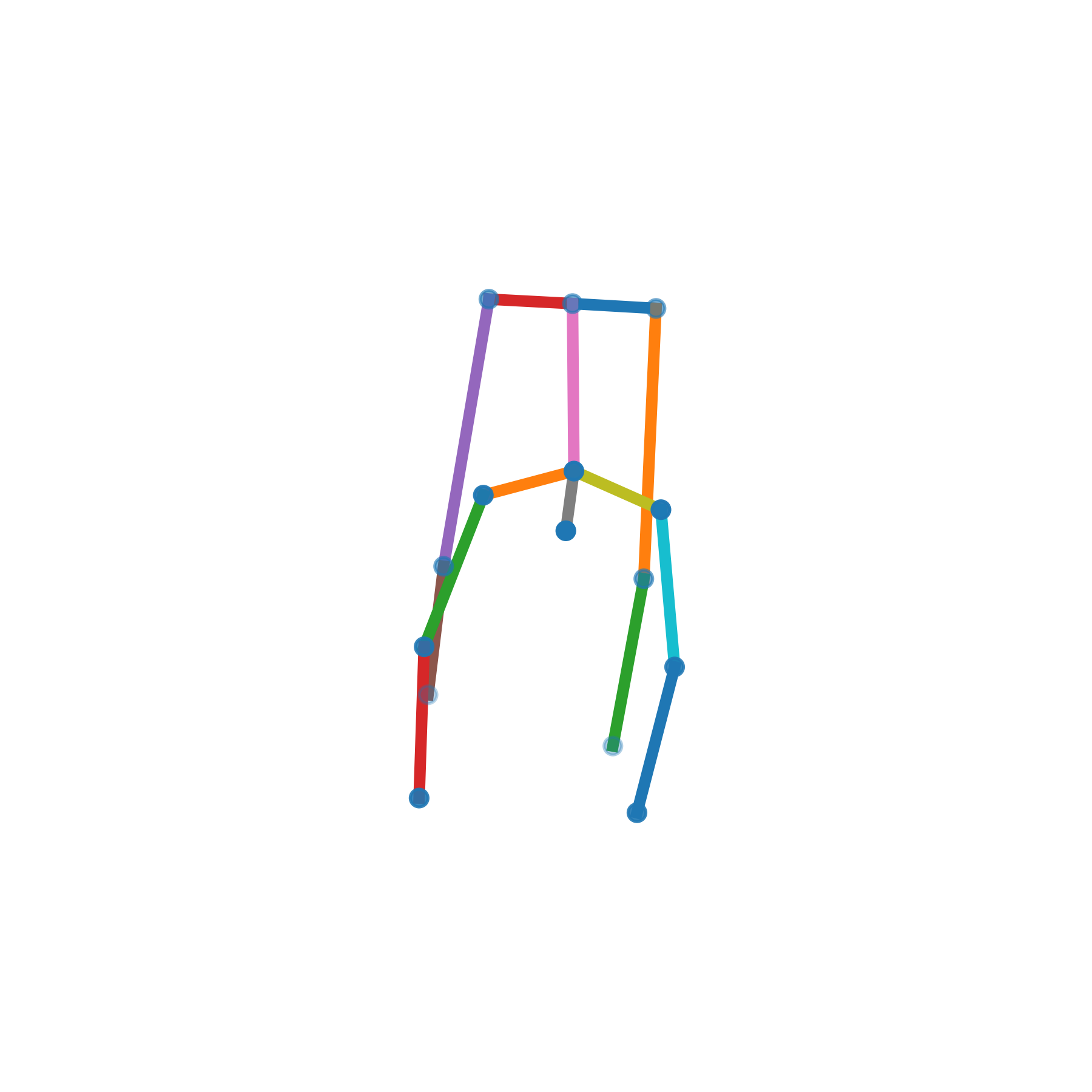}} &
        \makebox[\linewidth][c]{\includegraphics[width=0.82\linewidth]{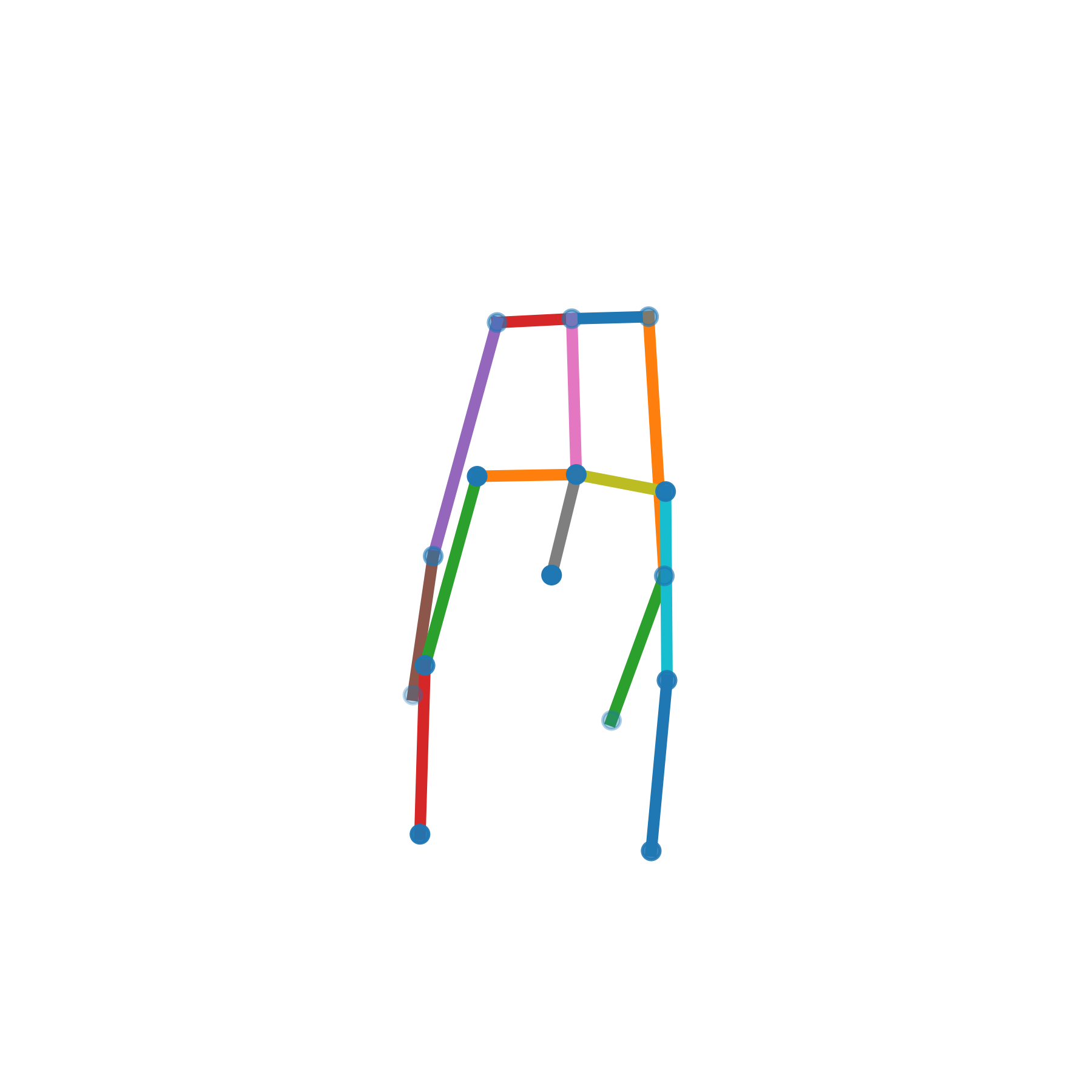}} &
        \makebox[\linewidth][c]{\includegraphics[width=0.82\linewidth]{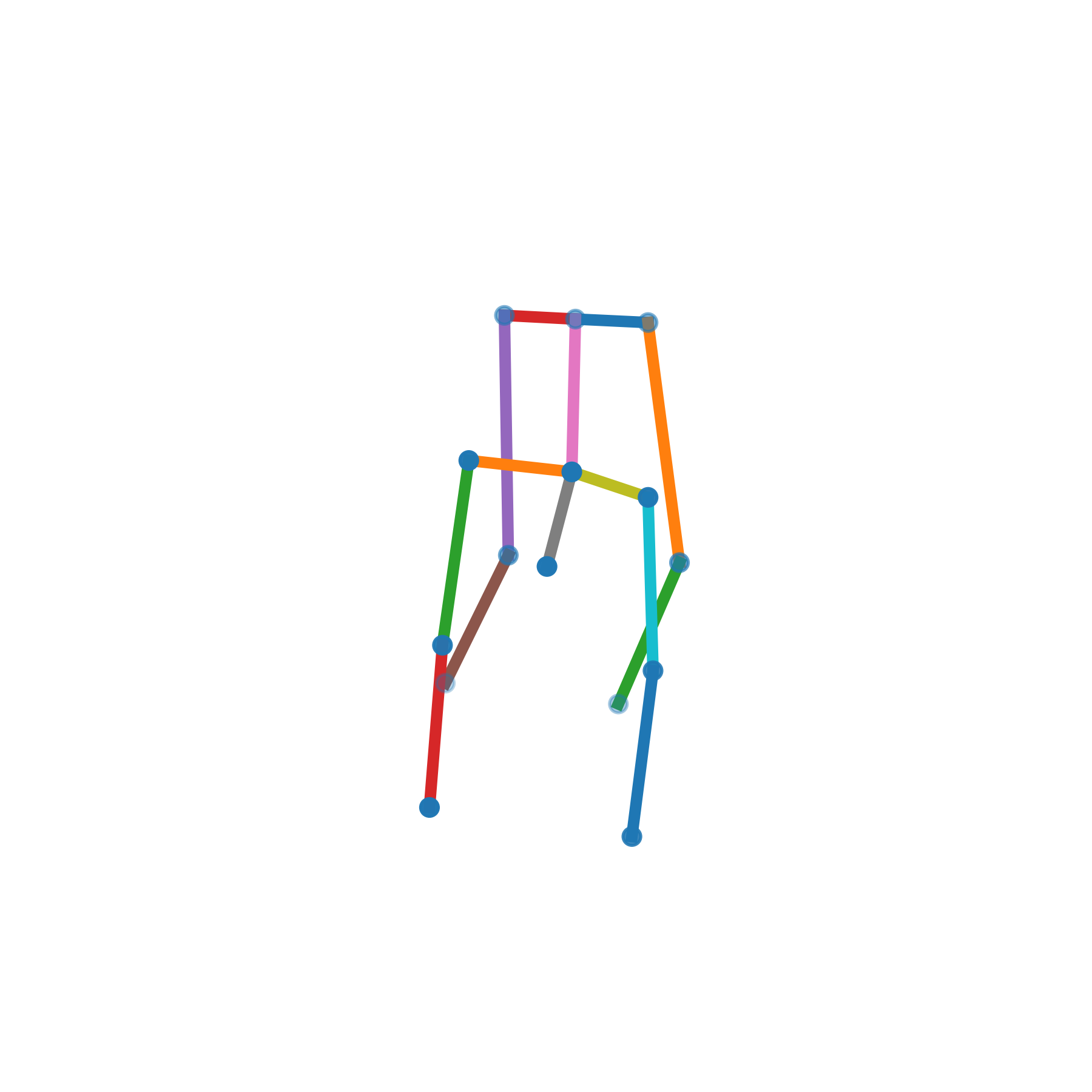}} &
        \makebox[\linewidth][c]{\includegraphics[width=0.82\linewidth]{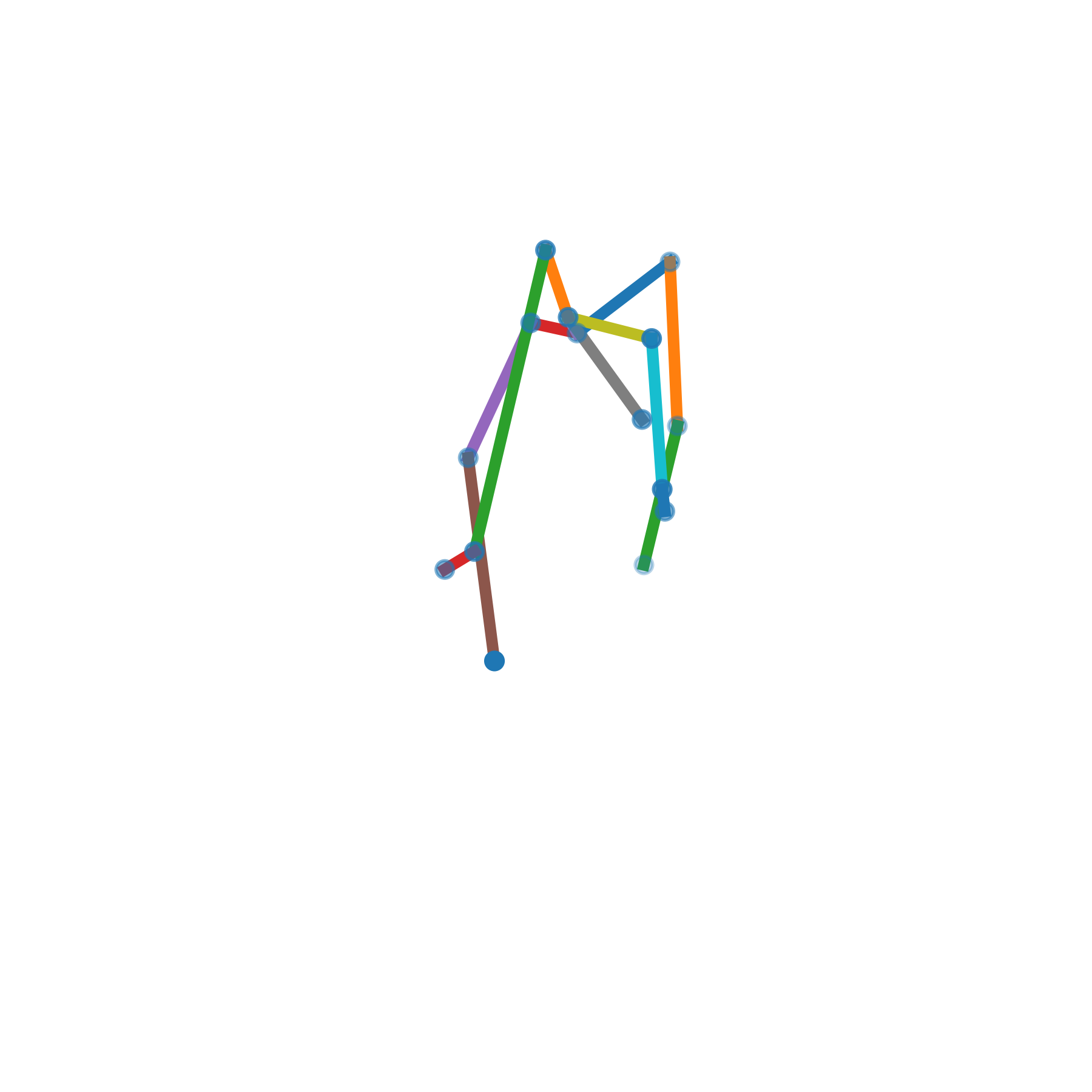}} \\[0.15cm]
        \textbf{1958} &
        \makebox[\linewidth][c]{\includegraphics[width=0.85\linewidth, trim=75 75 75 75, clip]{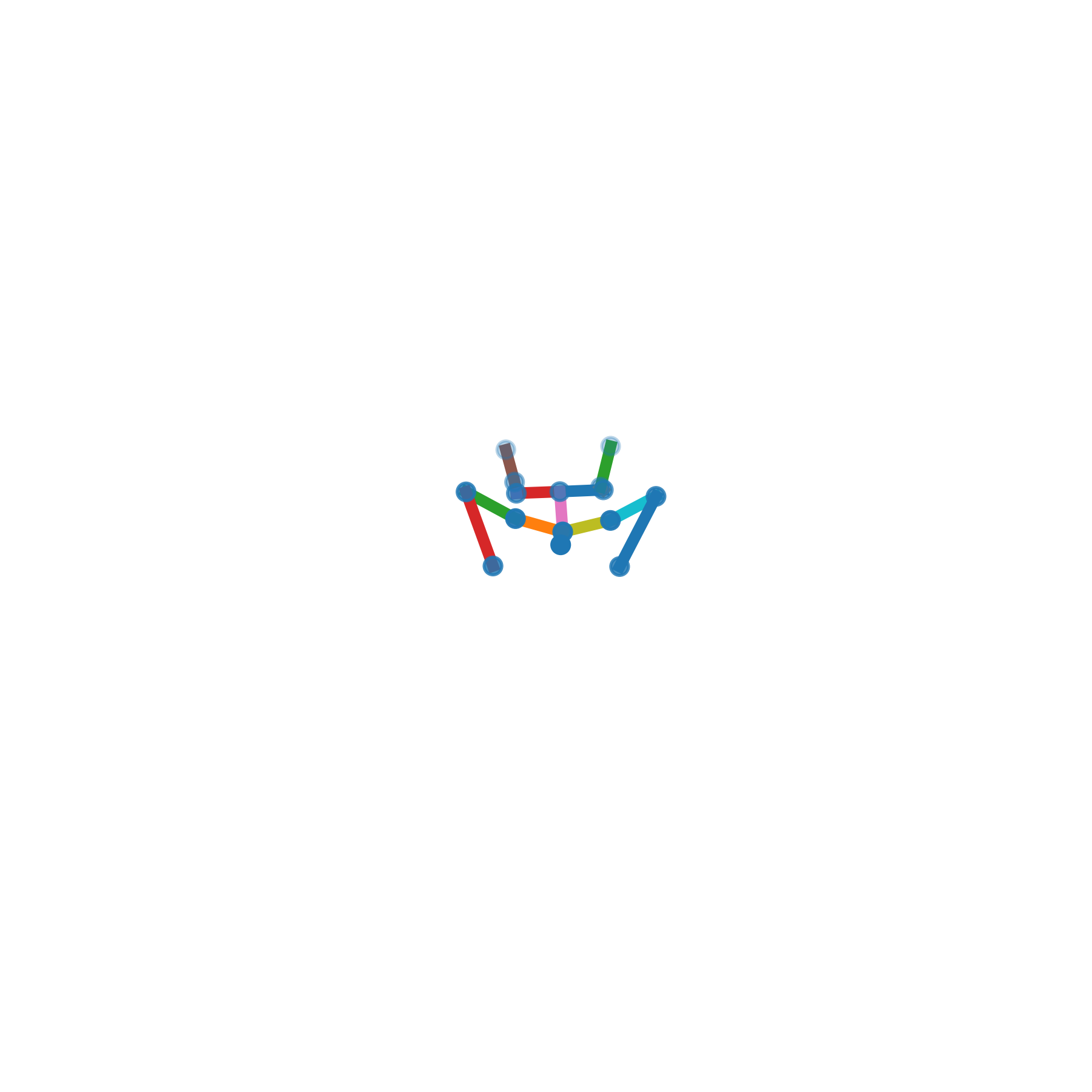}} &
        \makebox[\linewidth][c]{\includegraphics[width=0.85\linewidth, trim=75 75 75 75, clip]{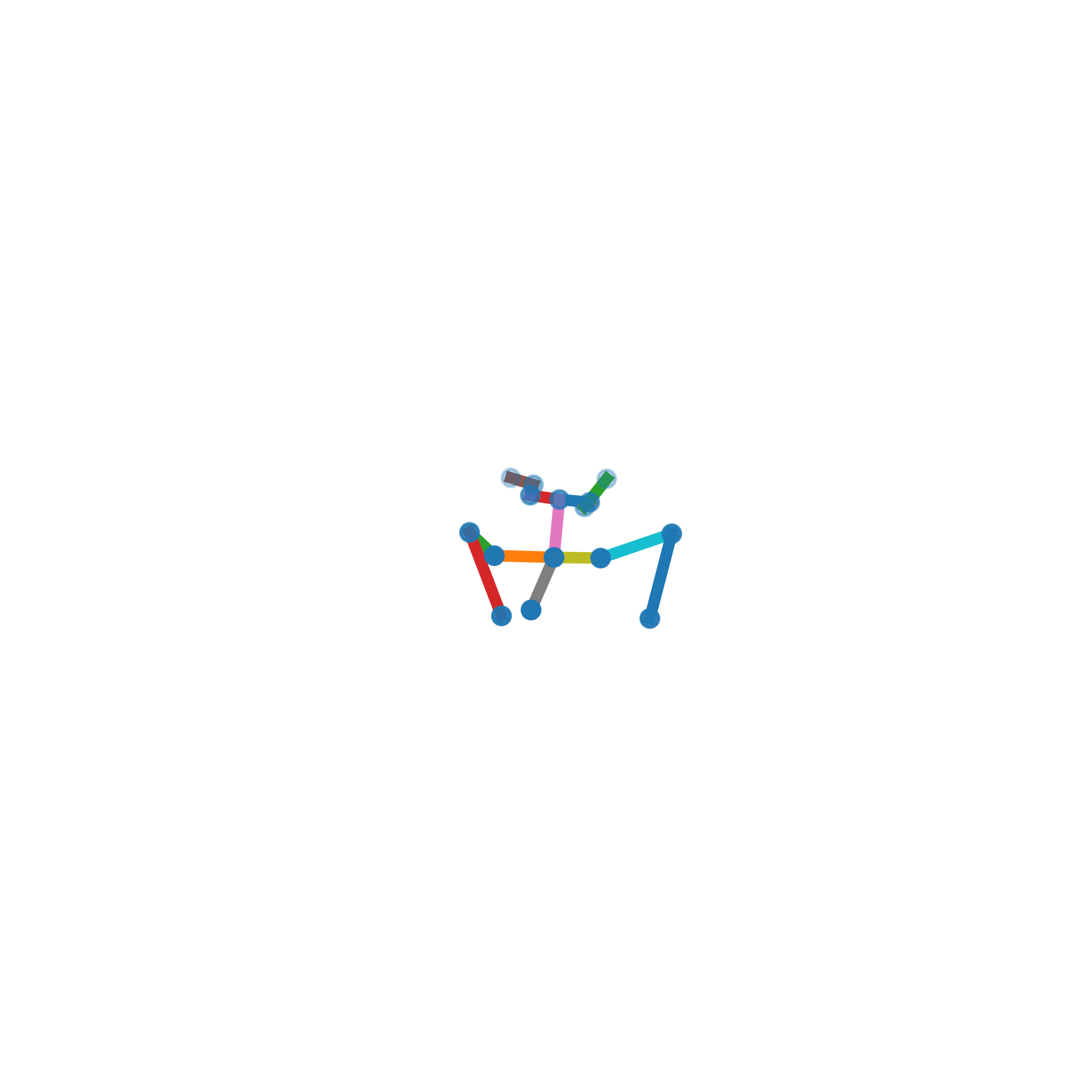}} &
        \makebox[\linewidth][c]{\includegraphics[width=0.85\linewidth, trim=75 75 75 75, clip]{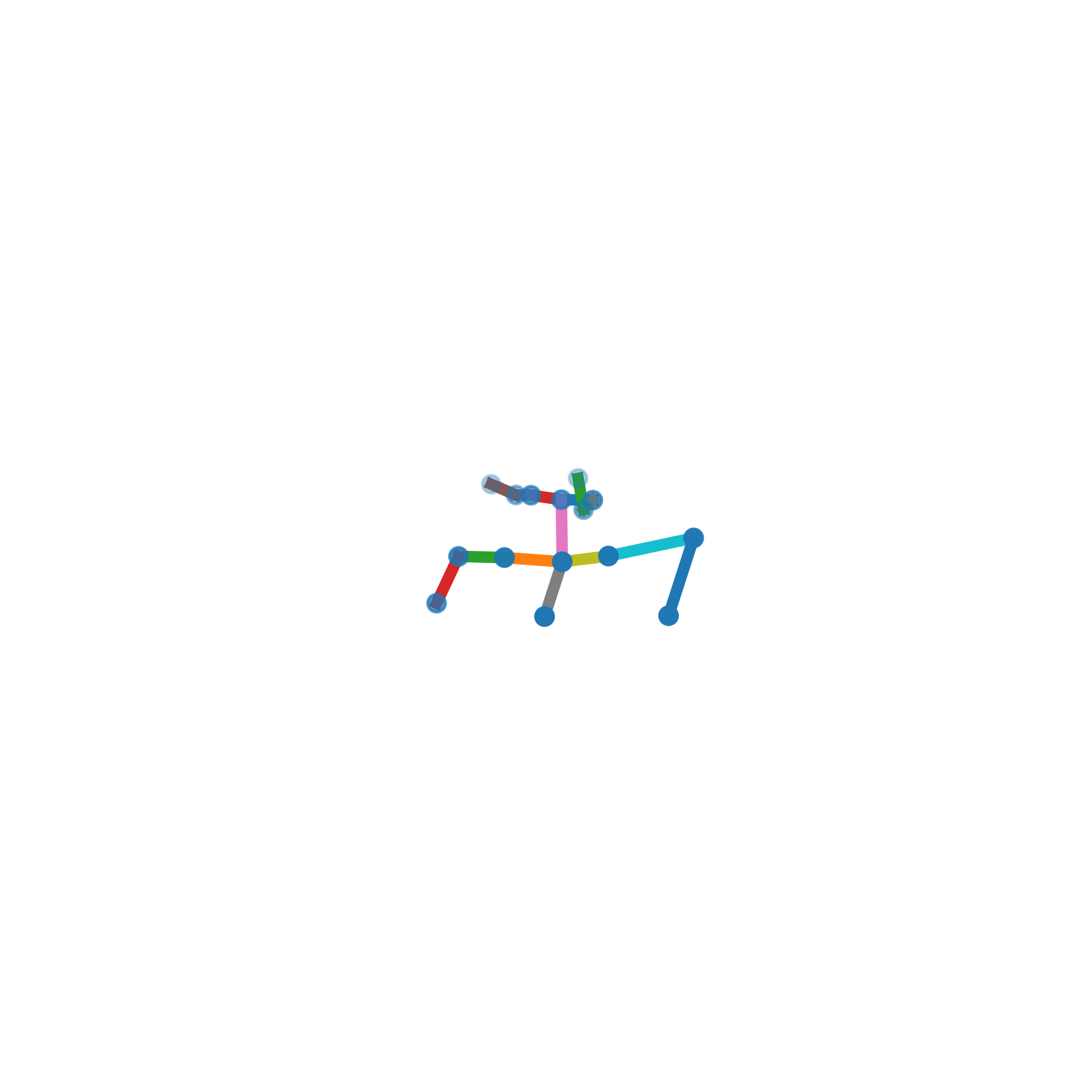}} &
        \makebox[\linewidth][c]{\includegraphics[width=0.85\linewidth, trim=75 75 75 75, clip]{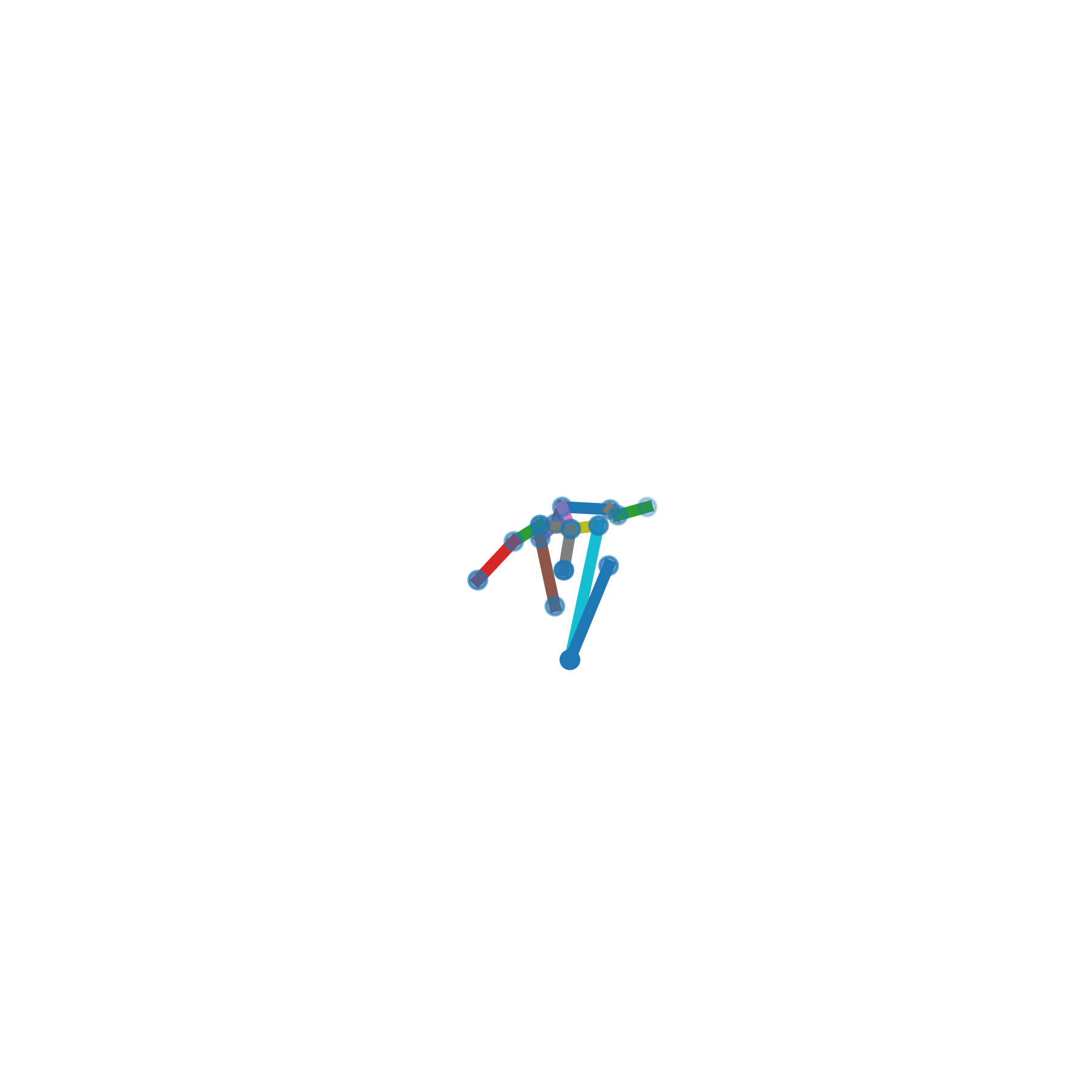}} \\
        \bottomrule
    \end{tabular}

    \caption{Qualitative comparison of 3D pose estimations for the \textit{man maker} sequence at three representative frames. Ground truth is compared with TCPFormer, PoseMamba-L, and MVGFormer predictions. In frames 1854 and 1958, the person lies on the floor, where the pose is more difficult to estimate, especially for MVGFormer.}
    \label{fig:man_maker_qualitative_comparison}
\end{figure}

Fig.~\ref{fig:comparison_exercise_error} complements the global comparison by showing how the error varies across exercises. The results show that the advantage of MVGFormer is not uniform across all actions. In relatively controlled standing exercises, the multi-view model usually yields lower errors. However, its performance degrades noticeably in more complex actions when the subject's body lies on the floor or adopts a strongly inclined posture, such as \textit{mule kick}, \textit{pushup}, \textit{diamond pushup}, \textit{burpees}, and \textit{man maker}. These exercises involve body configurations that differ considerably from the upright poses commonly represented in the multi-view training data, potentially contributing to the observed reconstruction difficulties. Fig.~\ref{fig:man_maker_qualitative_comparison} shows a qualitative comparison of the skeletons obtained by the three models for the \textit{man maker} sequence, illustrating the difficulties encountered by MVGFormer in this type of exercise.

This behavior can also be related to the internal pose representation used by MVGFormer. The model relies on compositional pose queries whose geometry terms are initialized from a canonical upright T-pose and then iteratively refined by the network. This initialization acts as a structural prior that encodes a standing body, with the limbs arranged around a vertical body axis. This prior is useful for many common standing poses; however, floor-supported exercises are much farther from this canonical configuration, and several joints can be strongly occluded or visually ambiguous. As a result, the model has to move away from its initial upright T-pose prior more aggressively, which can make the reconstruction less reliable. Additionally, MVGFormer performs the reconstruction frame by frame and does not explicitly exploit temporal information, so it cannot use motion continuity from adjacent frames to stabilize difficult poses or recover from temporary ambiguities.

\begin{figure}[!htbp]
    \centering
    \includegraphics[width=1.0\textwidth]{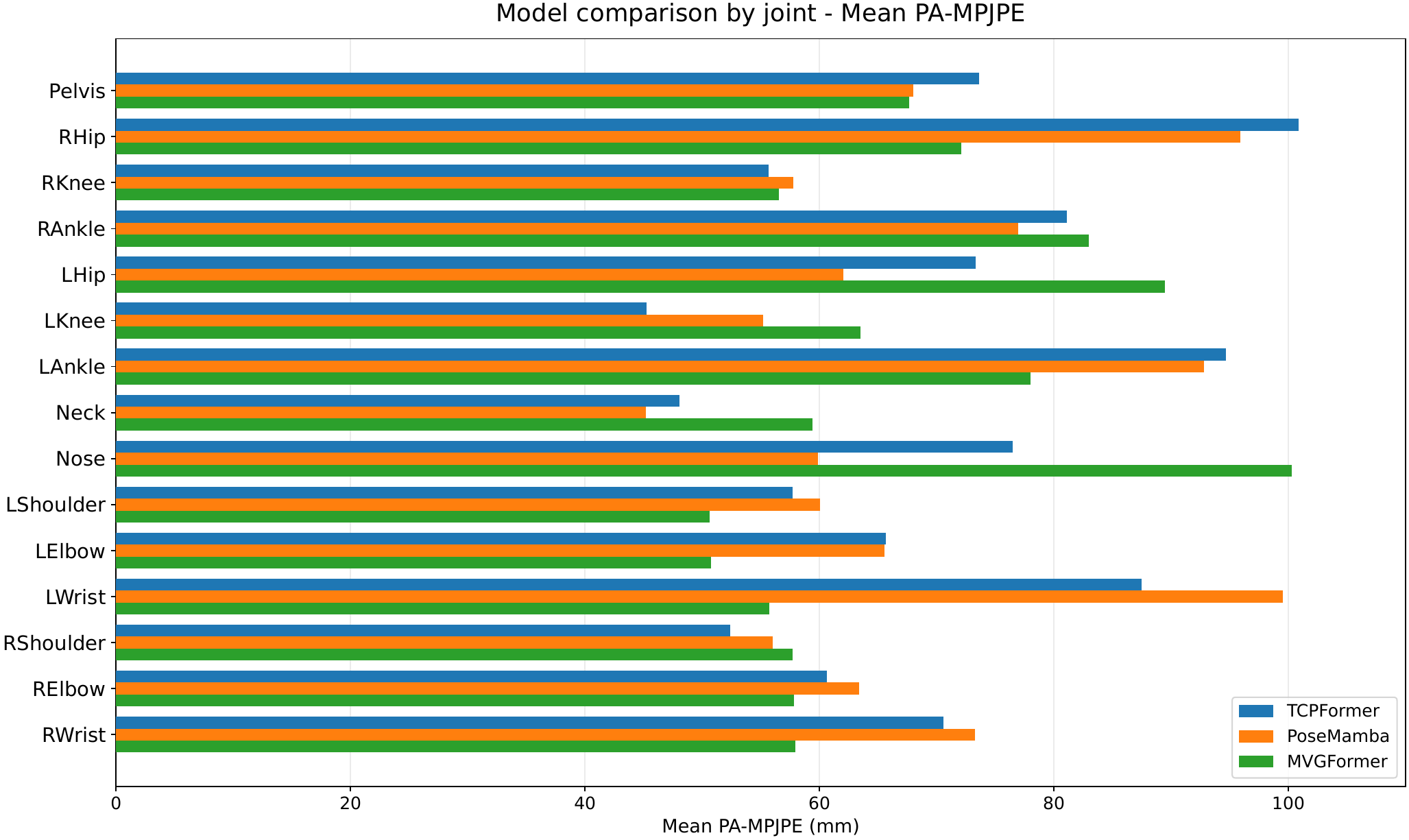}
    \caption{Mean PA-MPJPE per joint for TCPFormer, PoseMamba-L, and MVGFormer using the COMMON15 representation. For TCPFormer and PoseMamba-L, the Spine and Head joints are excluded from their original 17-joint representation, while the Nose joint is retained to match the corresponding joint predicted by MVGFormer.}
    \label{fig:comparison_joint_error}
\end{figure}

The per-joint comparison in Fig.~\ref{fig:comparison_joint_error} provides additional insight into the behavior of the evaluated models. TCPFormer and PoseMamba-L tend to show higher errors in distal joints, particularly the wrists and ankles, which are more mobile and can be more affected by occlusions or inaccuracies in the 2D detections. PoseMamba-L exhibits its highest error at the left wrist, while TCPFormer presents particularly high errors at the right hip and left ankle. In contrast, MVGFormer obtains lower errors for several limb joints, especially the wrists, but presents notably higher errors for the Nose and left hip. The hip joints also show relatively high errors for all models, which may partly reflect differences in the definition of these joints between the skeletal conventions. These differences highlight how the reconstruction error is distributed unevenly across body joints and depends on the characteristics of each estimation approach.

Overall, multi-view reconstruction provides the lowest average PA-MPJPE but performs poorly in some floor-supported exercises, whereas the monocular temporal models remain competitive and more consistent across types of movement despite relying on a single view and external 2D detections. These findings highlight both the benefits of multi-view observations and the importance of considering challenging body configurations when evaluating 3D pose estimation models.

\subsection{Effect of temporal filtering}

Table~\ref{tab:pampjpe_filtering} shows the effect of temporal filtering on the spatial accuracy of each model. For TCPFormer and PoseMamba-L, temporal post-processing produces only minor changes in PA-MPJPE. TCPFormer obtains 69.566\,mm for the original predictions, compared with 69.713\,mm using the One Euro Filter, 69.555\,mm using Savitzky--Golay, and 69.563\,mm using the Median Filter. PoseMamba-L is even less affected, with a mean PA-MPJPE ranging from 68.767 to 68.769\,mm across the original and filtered predictions. These results indicate that temporal filtering has a limited effect on the spatial accuracy of the monocular models, although the One Euro Filter introduces a slight increase in error for TCPFormer.

MVGFormer exhibits a more pronounced response to temporal filtering. Its original PA-MPJPE is 66.691\,mm, while the One Euro Filter slightly increases the error to 67.091\,mm. In contrast, Savitzky--Golay reduces PA-MPJPE to 62.102\,mm, and the Median Filter achieves the lowest value at 60.385\,mm. Because MVGFormer reconstructs each frame independently, isolated reconstruction errors may produce spatial outliers and abrupt variations in the predicted joint trajectories. Temporal filtering can attenuate these irregularities, improving temporal consistency and, in the case of Savitzky--Golay and Median filtering, also reducing the average spatial error.

\begin{table}[!htbp]
\centering
\caption{Comparison of mean PA-MPJPE (mm) per subject and model, before and after temporal filtering, evaluated on the common 15-joint skeleton. TCPFormer and PoseMamba-L use RTMPose as the 2D input source.}
\label{tab:pampjpe_filtering}
\resizebox{\linewidth}{!}{%
\begin{tabular}{lccccccccc}
\toprule
Model / version & s03 & s04 & s05 & s07 & s08 & s09 & s10 & s11 & \textbf{Mean} \\
\midrule

TCPFormer -- Original
& 68.394 & 71.662 & 71.473 & 70.798 & 72.220 & 70.441 & 67.010 & 64.527
& \textbf{69.566} \\

TCPFormer -- One Euro filter
& 68.560 & 71.779 & 71.621 & 70.897 & 72.427 & 70.552 & 67.194 & 64.679
& \textbf{69.713} \\

TCPFormer -- Savitzky--Golay filter
& 68.386 & 71.651 & 71.460 & 70.787 & 72.212 & 70.431 & 67.001 & 64.511
& \textbf{69.555} \\

TCPFormer -- Median filter
& 68.394 & 71.661 & 71.470 & 70.794 & 72.220 & 70.439 & 67.008 & 64.517
& \textbf{69.563} \\

\addlinespace

PoseMamba-L -- Original
& 68.922 & 74.660 & 68.557 & 68.321 & 71.521 & 69.209 & 64.481 & 64.476
& \textbf{68.768} \\

PoseMamba-L -- One Euro filter
& 68.923 & 74.656 & 68.555 & 68.318 & 71.526 & 69.207 & 64.485 & 64.478
& \textbf{68.769} \\

PoseMamba-L -- Savitzky--Golay filter
& 68.922 & 74.654 & 68.557 & 68.321 & 71.521 & 69.209 & 64.481 & 64.475
& \textbf{68.767} \\

PoseMamba-L -- Median filter
& 68.923 & 74.654 & 68.557 & 68.321 & 71.521 & 69.209 & 64.481 & 64.475
& \textbf{68.768} \\

\addlinespace

MVGFormer -- Original
& 61.672 & 60.299 & 67.718 & 70.701 & 77.136 & 71.240 & 55.807 & 68.952
& \textbf{66.691} \\

MVGFormer -- One Euro filter
& 63.150 & 60.748 & 67.726 & 70.766 & 76.539 & 71.101 & 58.285 & 68.415
& \textbf{67.091} \\

MVGFormer -- Savitzky--Golay filter
& 56.937 & 56.006 & 62.148 & 66.700 & 72.272 & 66.835 & 52.274 & 63.644
& \textbf{62.102} \\

MVGFormer -- Median filter
& 55.898 & 55.140 & 59.776 & 65.407 & 70.535 & 63.953 & 50.397 & 61.973
& \textbf{60.385} \\

\bottomrule
\end{tabular}}
\end{table}

However, spatial accuracy alone is not sufficient to evaluate 3D pose estimation in video sequences. A model may obtain competitive PA-MPJPE values while still producing unstable trajectories characterized by frame-to-frame jitter or abrupt joint displacements. Therefore, we use MPJVE to evaluate temporal consistency.

\begin{table}[!htbp]
\centering
\caption{Mean Per-Joint Velocity Error (MPJVE) with respect to the ground truth (mm/frame) per subject, comparing the original predictions with the One Euro, Savitzky--Golay, and Median filters, evaluated on the common 15-joint skeleton. TCPFormer and PoseMamba-L use RTMPose as the 2D input source.}
\label{tab:mpjve}
\resizebox{\linewidth}{!}{%
\begin{tabular}{lccccccccc}
\toprule
Model / version & s03 & s04 & s05 & s07 & s08 & s09 & s10 & s11 & \textbf{Mean} \\
\midrule
TCPFormer -- Original                 & 2.823 & 2.867 & 3.216 & 3.047 & 3.325 & 2.685 & 2.925 & 2.659 & \textbf{2.943} \\
TCPFormer -- One Euro filter          & 2.803 & 2.848 & 3.187 & 3.011 & 3.323 & 2.632 & 2.887 & 2.608 & \textbf{2.913} \\
TCPFormer -- Savitzky--Golay filter   & 2.764 & 2.825 & 3.149 & 2.972 & 3.285 & 2.592 & 2.847 & 2.557 & \textbf{2.874} \\
TCPFormer -- Median filter            & 2.792 & 2.848 & 3.184 & 3.007 & 3.308 & 2.627 & 2.880 & 2.593 & \textbf{2.905} \\
\addlinespace
PoseMamba-L -- Original               & 2.774 & 2.824 & 3.136 & 2.938 & 3.240 & 2.578 & 2.841 & 2.539 & \textbf{2.859} \\
PoseMamba-L -- One Euro filter        & 2.775 & 2.824 & 3.136 & 2.938 & 3.241 & 2.578 & 2.841 & 2.539 & \textbf{2.859} \\
PoseMamba-L -- Savitzky--Golay filter & 2.771 & 2.821 & 3.133 & 2.936 & 3.238 & 2.575 & 2.838 & 2.536 & \textbf{2.856} \\
PoseMamba-L -- Median filter          & 2.774 & 2.823 & 3.136 & 2.938 & 3.240 & 2.578 & 2.840 & 2.538 & \textbf{2.858} \\
\addlinespace
MVGFormer -- Original                 & 26.014 & 23.492 & 28.963 & 26.565 & 27.800 & 27.443 & 23.059 & 28.332 & \textbf{26.458} \\
MVGFormer -- One Euro filter          & 6.646 & 5.769 & 7.449 & 6.277 & 7.081 & 6.521 & 5.552 & 7.103 & \textbf{6.550} \\
MVGFormer -- Savitzky--Golay filter   & 4.696 & 4.407 & 5.393 & 4.910 & 5.509 & 4.872 & 4.228 & 5.228 & \textbf{4.905} \\
MVGFormer -- Median filter            & 4.056 & 3.860 & 4.642 & 4.326 & 4.708 & 3.950 & 3.881 & 4.116 & \textbf{4.192} \\
\bottomrule
\end{tabular}}
\end{table}

Table~\ref{tab:mpjve} reports the MPJVE before and after temporal post-processing. The original predictions of TCPFormer and PoseMamba-L exhibit much lower temporal errors than MVGFormer, with mean values of 2.94 and 2.86\,mm/frame, respectively. This is consistent with the explicit temporal modeling employed by both monocular architectures. For TCPFormer, all three filters reduce MPJVE: the One Euro Filter decreases it to 2.91\,mm/frame, the Median Filter to 2.90\,mm/frame, and Savitzky--Golay achieves the lowest value at 2.87\,mm/frame, representing a reduction of approximately 2.34\% compared with the original predictions. In contrast, PoseMamba-L shows a much smaller margin for improvement. These results indicate that TCPFormer benefits slightly from additional temporal smoothing, whereas the improvements obtained by PoseMamba-L are negligible.

The behavior of MVGFormer is markedly different. Its unfiltered predictions exhibit a substantially higher MPJVE of 26.46\,mm/frame, consistent with its frame-wise reconstruction. In this case, temporal post-processing produces a much larger improvement. The One Euro Filter reduces MPJVE to 6.55\,mm/frame, Savitzky--Golay further reduces it to 4.91\,mm/frame, and the Median Filter achieves the lowest value at 4.19\,mm/frame. This corresponds to a reduction of approximately 84.2\% relative to the original predictions. Such a substantial improvement suggests that a large part of the temporal error in MVGFormer is associated with abrupt frame-to-frame variations that can be effectively attenuated through temporal filtering.

Considering Tables~\ref{tab:pampjpe_filtering} and~\ref{tab:mpjve} jointly, Savitzky--Golay is the best option for both monocular models, reducing MPJVE without affecting PA-MPJPE, although the gain is marginal for PoseMamba-L. For MVGFormer, the Median Filter provides the best overall result, improving both metrics simultaneously. The effectiveness of temporal post-processing therefore depends strongly on the temporal characteristics of the underlying model.

To complement the MPJVE results, we inspect the mean body velocity defined in (\ref{eq:velocity}), which helps identify abrupt peaks, persistent jitter, temporal delays, and changes in the motion profile introduced by the filters. We selected the \textit{band pull apart} sequence of subject s03 because its clear and repetitive movement pattern facilitates the comparison.

\begin{figure}[!htbp]
    \centering
    \begin{tabular}{@{}c@{}}
        \includegraphics[width=1.0\textwidth]{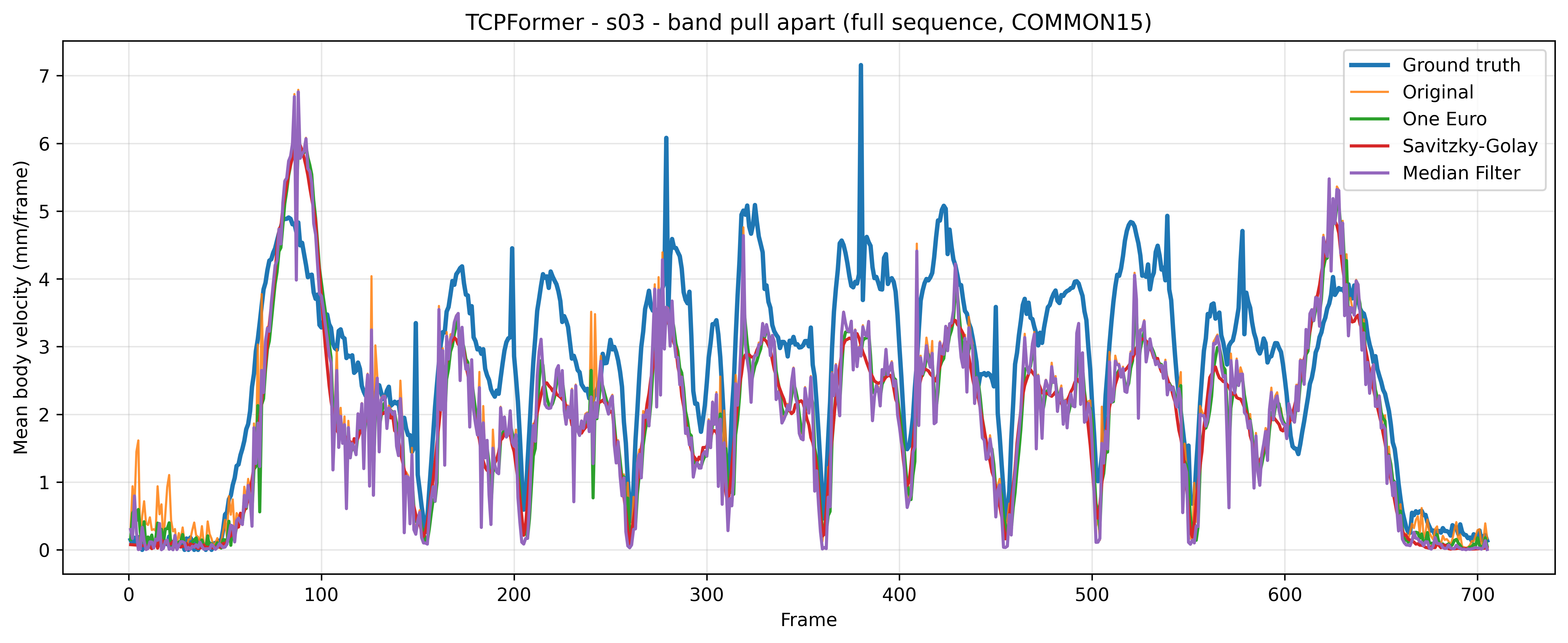} \\[0.3cm]
        \includegraphics[width=1.0\textwidth]{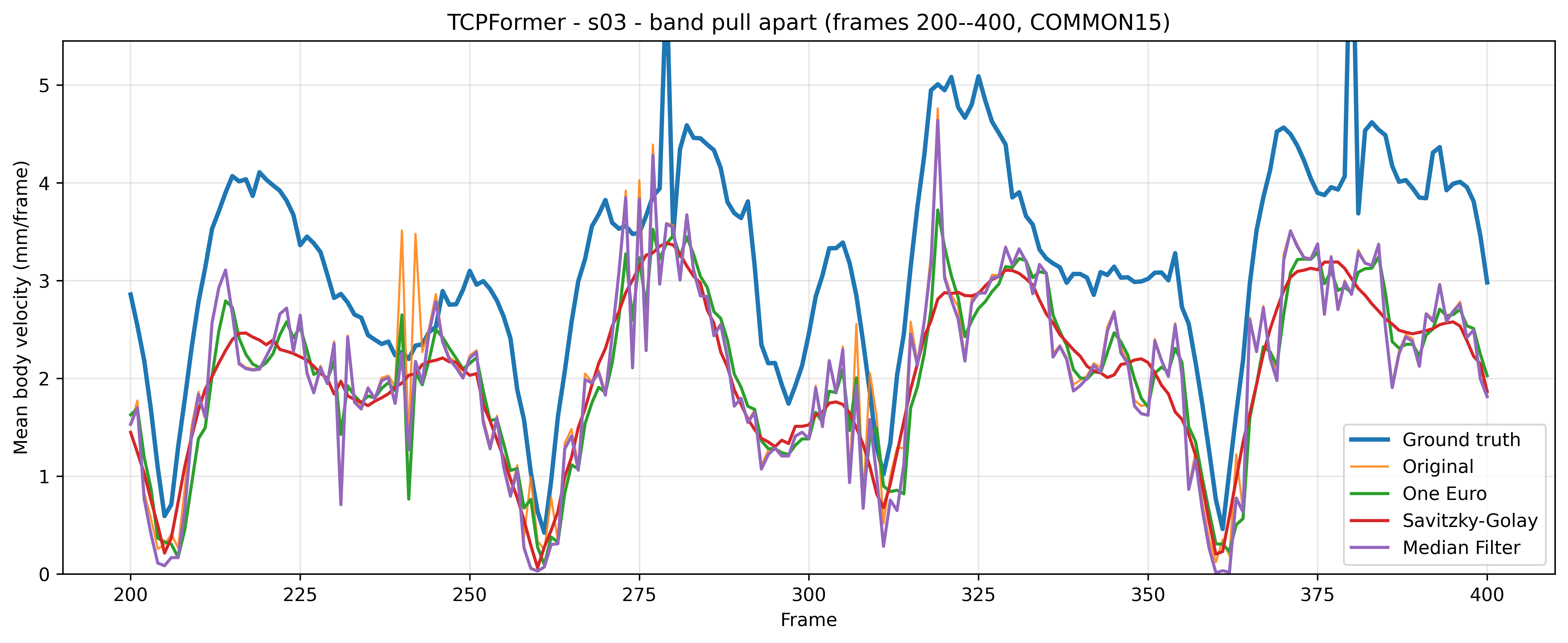}
    \end{tabular}
    \caption{Mean body velocity for TCPFormer on the \textit{band pull apart} sequence of subject s03, evaluated using COMMON15. The top plot shows the complete sequence, while the bottom plot shows a zoomed view between frames 200 and 400, comparing the ground truth, original predictions, and temporally filtered trajectories.}
    \label{fig:tcpformer_velocity_filters}
\end{figure}

\begin{figure}[!htbp]
    \centering
    \begin{tabular}{@{}c@{}}
        \includegraphics[width=1.0\textwidth]{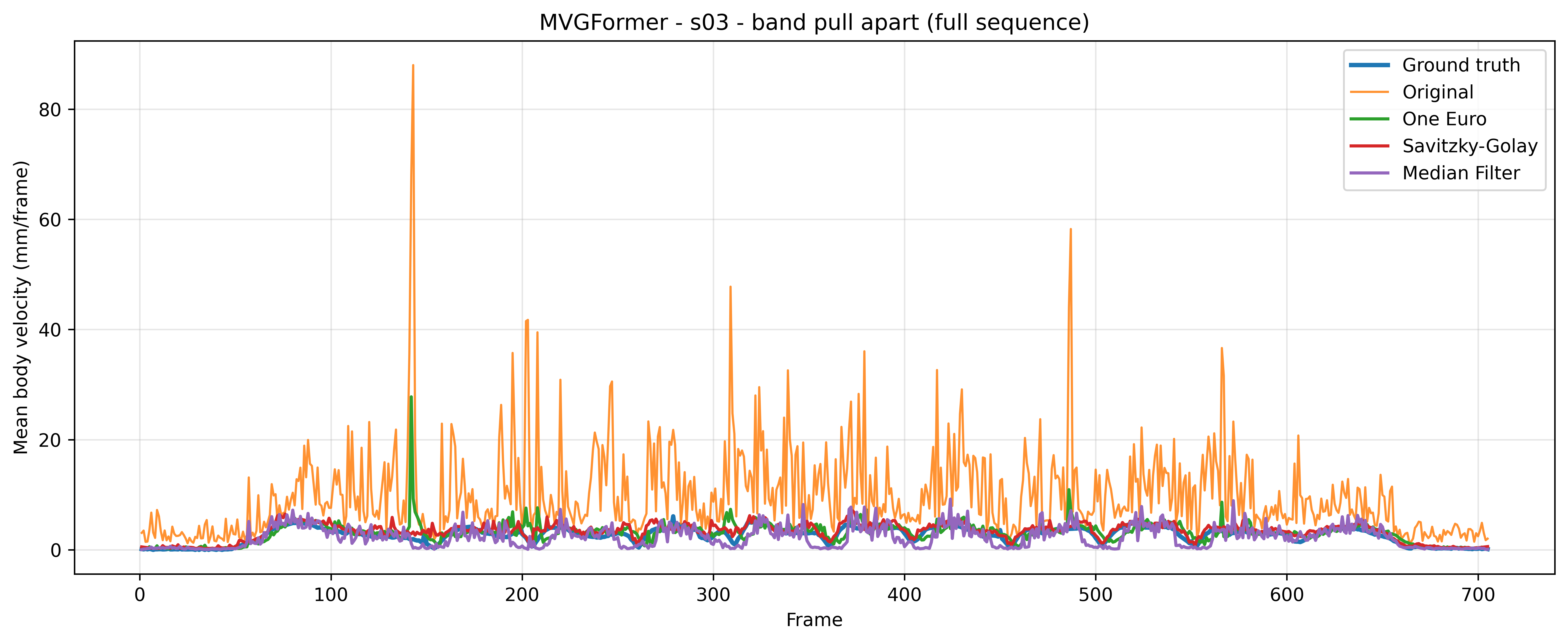} \\[0.3cm]
        \includegraphics[width=1.0\textwidth]{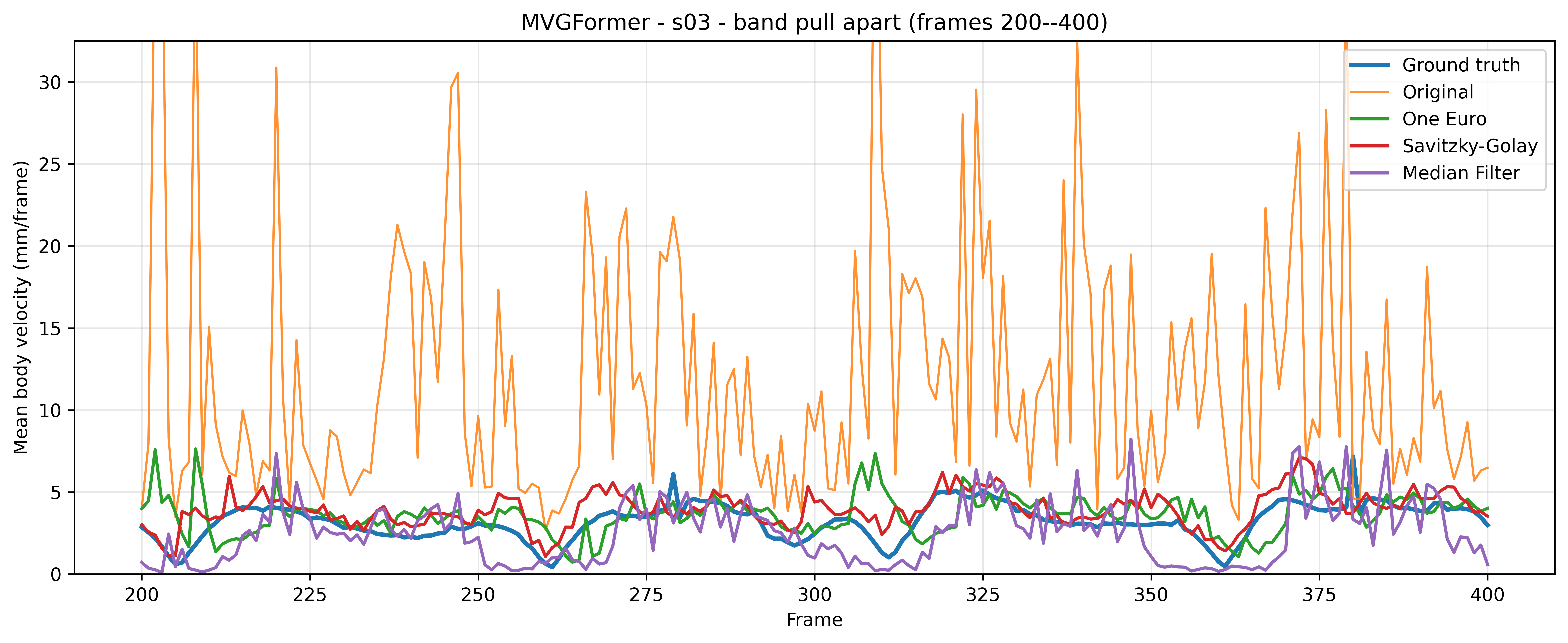}
    \end{tabular}
    \caption{Mean body velocity for MVGFormer on the \textit{band pull apart} sequence of subject s03, evaluated using COMMON15. The top plot shows the complete sequence, while the bottom plot shows a zoomed view between frames 200 and 400.}
    \label{fig:mvgformer_velocity_filters}
\end{figure}

The velocity curves in Figs.~\ref{fig:tcpformer_velocity_filters} and~\ref{fig:mvgformer_velocity_filters} compare, for TCPFormer and MVGFormer, the ground-truth velocity, the original prediction, and the three filtered versions, both over the full sequence and in a zoomed view of frames 200--400 that highlights local variations.

For TCPFormer, the original prediction reproduces the main temporal structure of the ground-truth motion, with the principal movement phases occurring at similar positions in the sequence. However, the predicted velocity profile generally exhibits a lower amplitude than the ground truth, indicating that some rapid changes in joint motion are underestimated. Local high-frequency variations are also present, although they are considerably smaller than those observed for MVGFormer.

The zoomed TCPFormer plot shows that temporal filtering mainly affects local fluctuations in the predicted velocity. Savitzky--Golay produces a more regular velocity profile while preserving the principal movement phases, which is consistent with its lowest MPJVE for this model. The One Euro Filter attenuates some abrupt variations while remaining relatively close to the original trajectory, whereas the Median Filter suppresses certain isolated peaks but retains noticeable short-term fluctuations. Nevertheless, the predicted velocities generally remain below the ground-truth values during several movement phases. Overall, temporal filtering modifies an already relatively stable trajectory, but does not fully correct the differences between the predicted and ground-truth motion.

MVGFormer exhibits a markedly different temporal behavior. The unfiltered prediction contains frequent high-amplitude frame-to-frame variations and isolated peaks, reflecting both persistent jitter and occasional reconstruction outliers. These irregularities are substantially larger than those observed in the ground-truth signal and are consistent with the high MPJVE of the original MVGFormer predictions.

The zoomed MVGFormer plot shows that all three filters substantially reduce these abrupt variations. The One Euro Filter strongly attenuates the velocity signal, whereas Savitzky--Golay provides a smoother trajectory while retaining more of the local motion structure. The Median Filter is particularly effective at suppressing isolated extreme variations and achieves the lowest MPJVE for MVGFormer. Therefore, although Savitzky--Golay may provide a visually smoother profile in some intervals, the quantitative results indicate that the Median Filter offers the best overall temporal agreement with the ground truth for this model.

\begin{figure}[!htbp]
    \centering
    \renewcommand{\arraystretch}{1.2}
    \setlength{\tabcolsep}{3pt}

    \begin{tabular}{
        >{\centering\arraybackslash}m{0.19\linewidth}
        >{\centering\arraybackslash}m{0.19\linewidth}
        >{\centering\arraybackslash}m{0.19\linewidth}
        >{\centering\arraybackslash}m{0.19\linewidth}
        >{\centering\arraybackslash}m{0.19\linewidth}
    }
        \toprule
        \textbf{Ground truth} & \textbf{MVGFormer} & \textbf{One Euro} & \textbf{Savitzky--Golay} & \textbf{Median Filter} \\
        \midrule

        \multicolumn{5}{c}{\textbf{\textit{band pull apart} -- Frame 142}} \\[0.1cm]

        \makebox[\linewidth][c]{\includegraphics[width=0.95\linewidth]{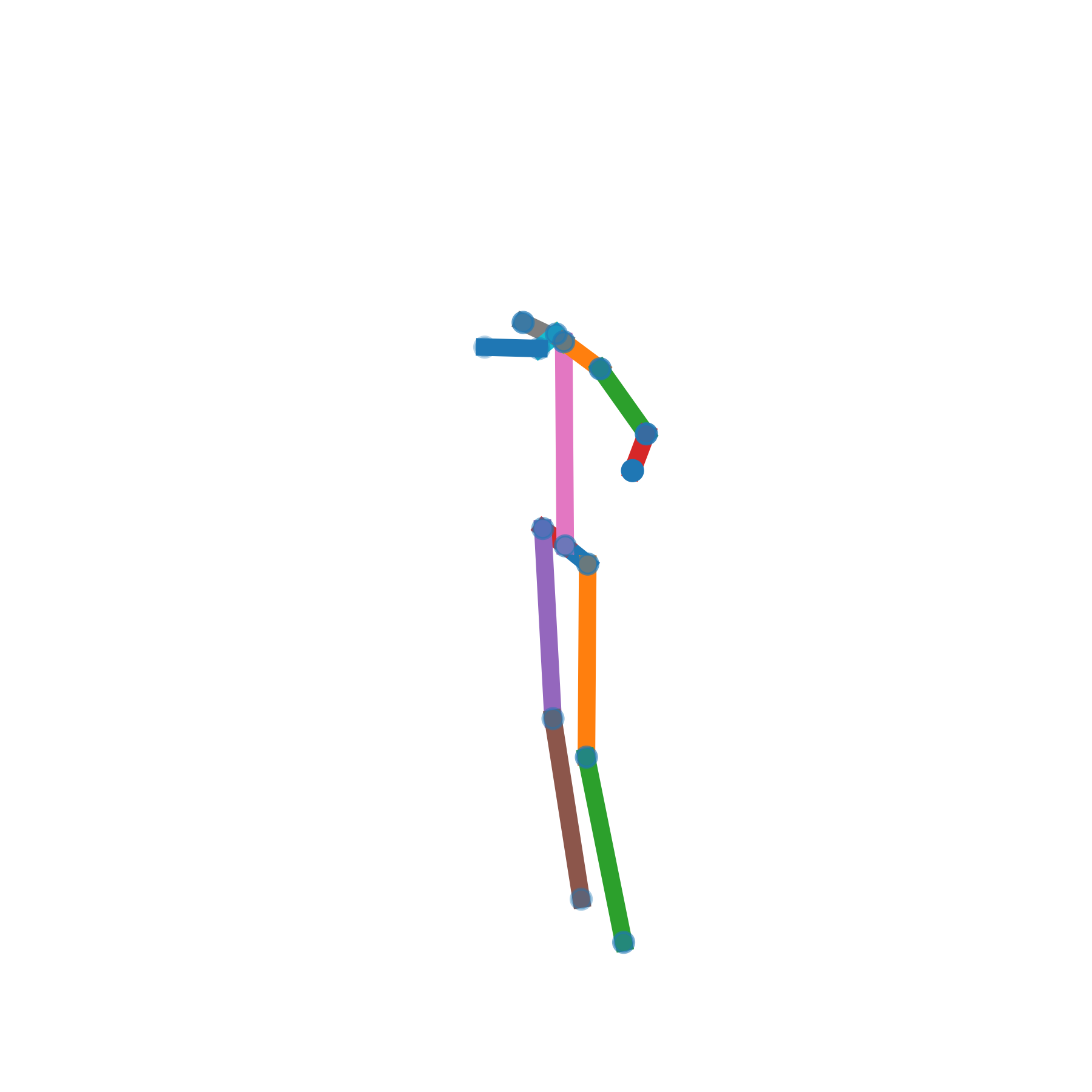}} &
        \makebox[\linewidth][c]{\includegraphics[width=0.95\linewidth]{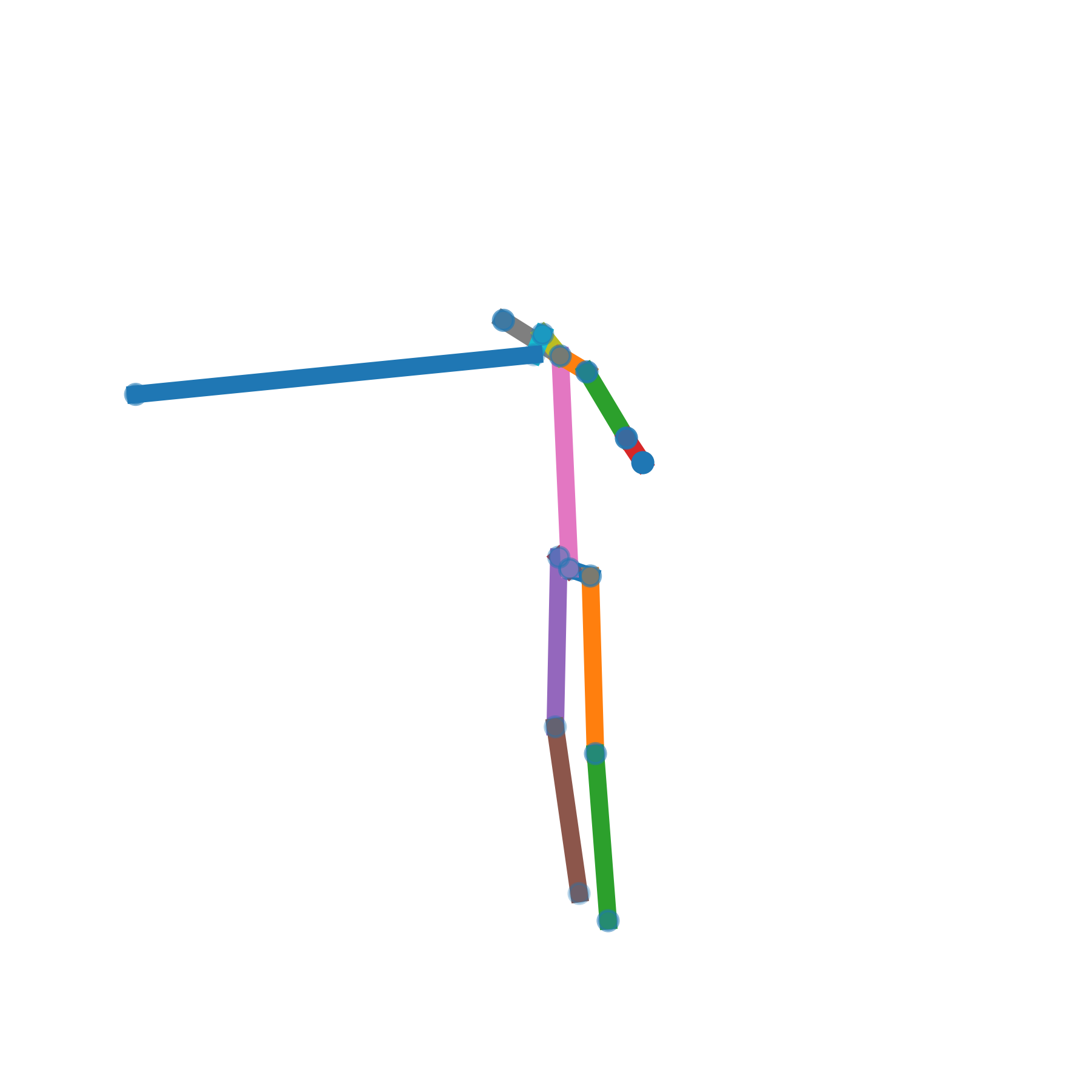}} &
        \makebox[\linewidth][c]{\includegraphics[width=0.95\linewidth]{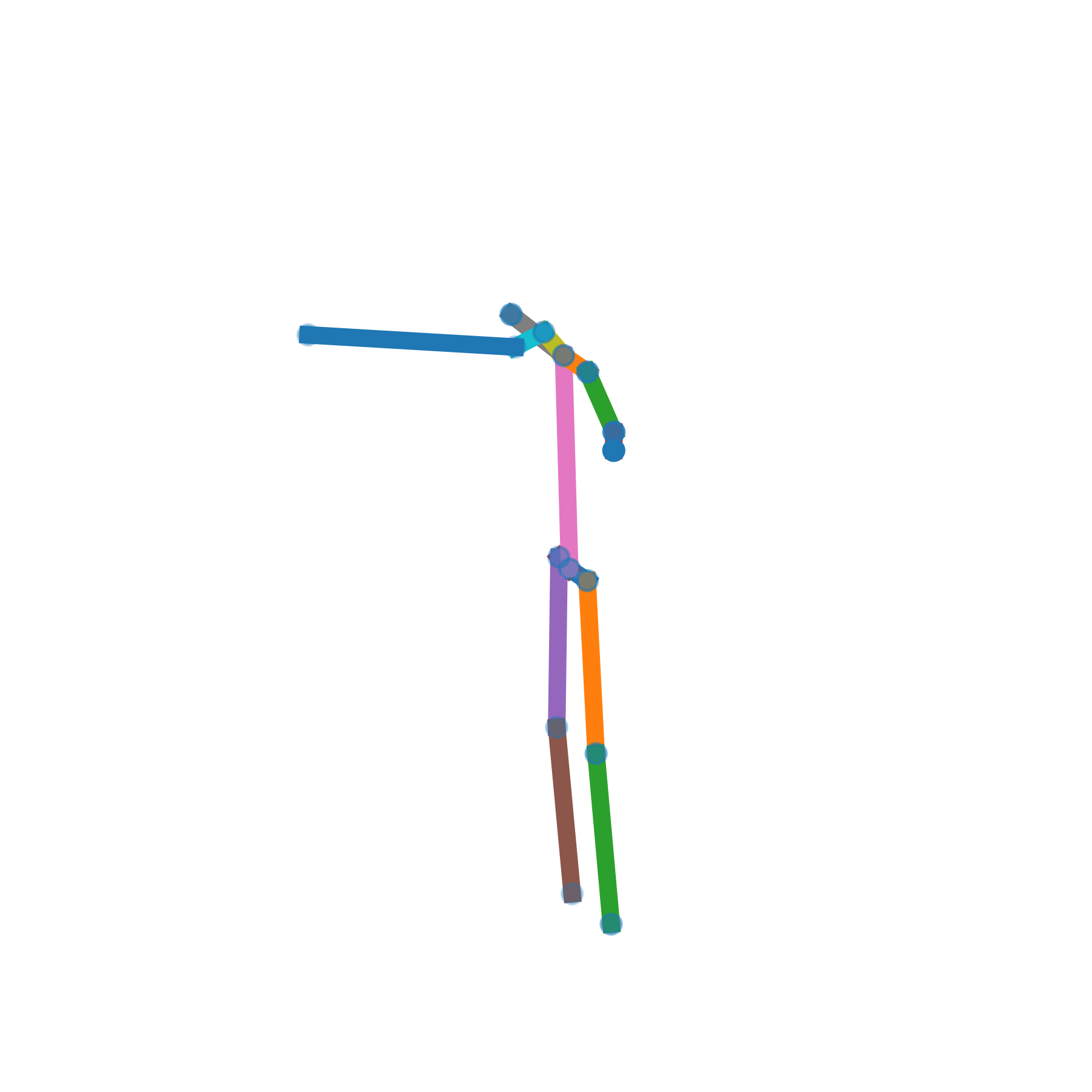}} &
        \makebox[\linewidth][c]{\includegraphics[width=0.95\linewidth]{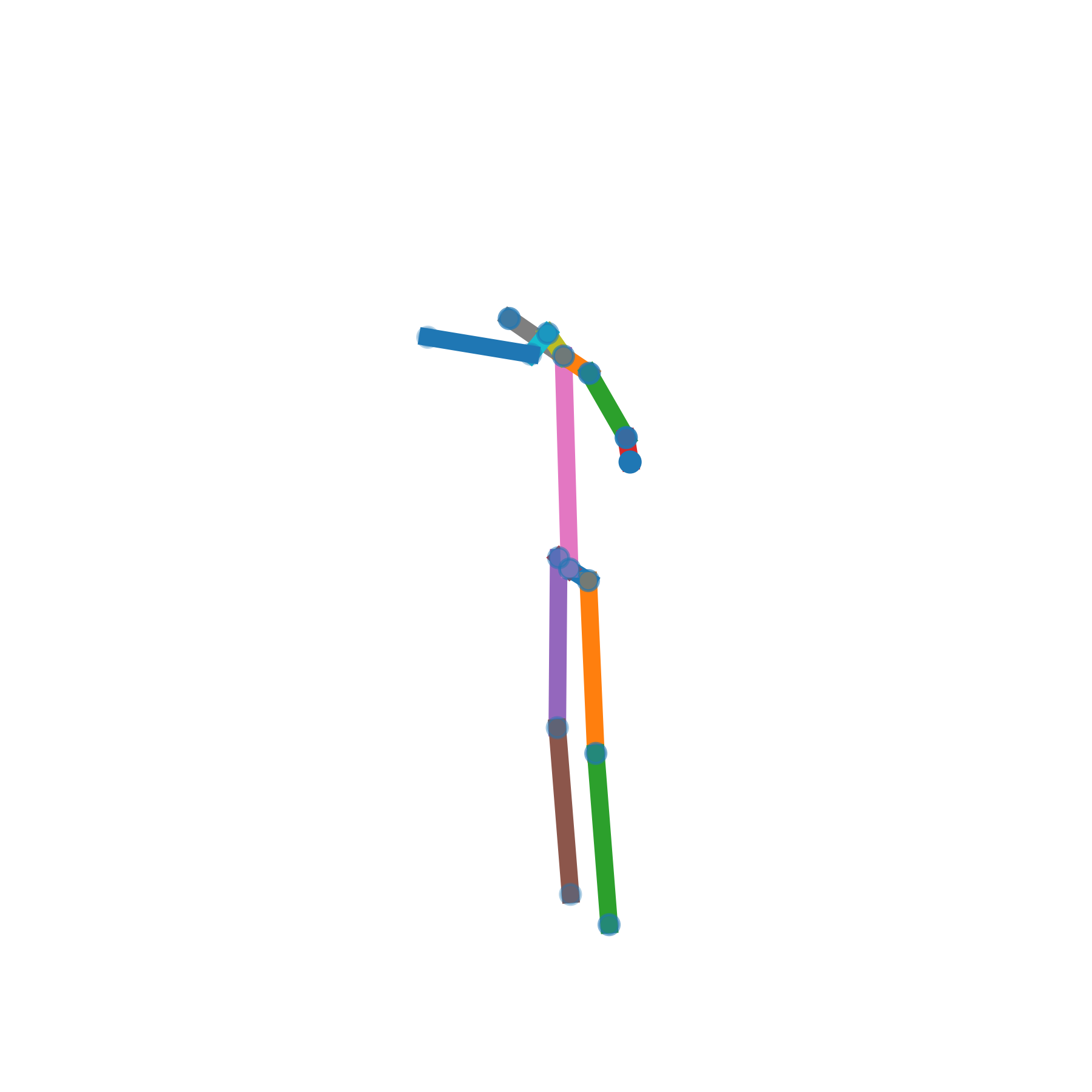}} &
        \makebox[\linewidth][c]{\includegraphics[width=0.95\linewidth]{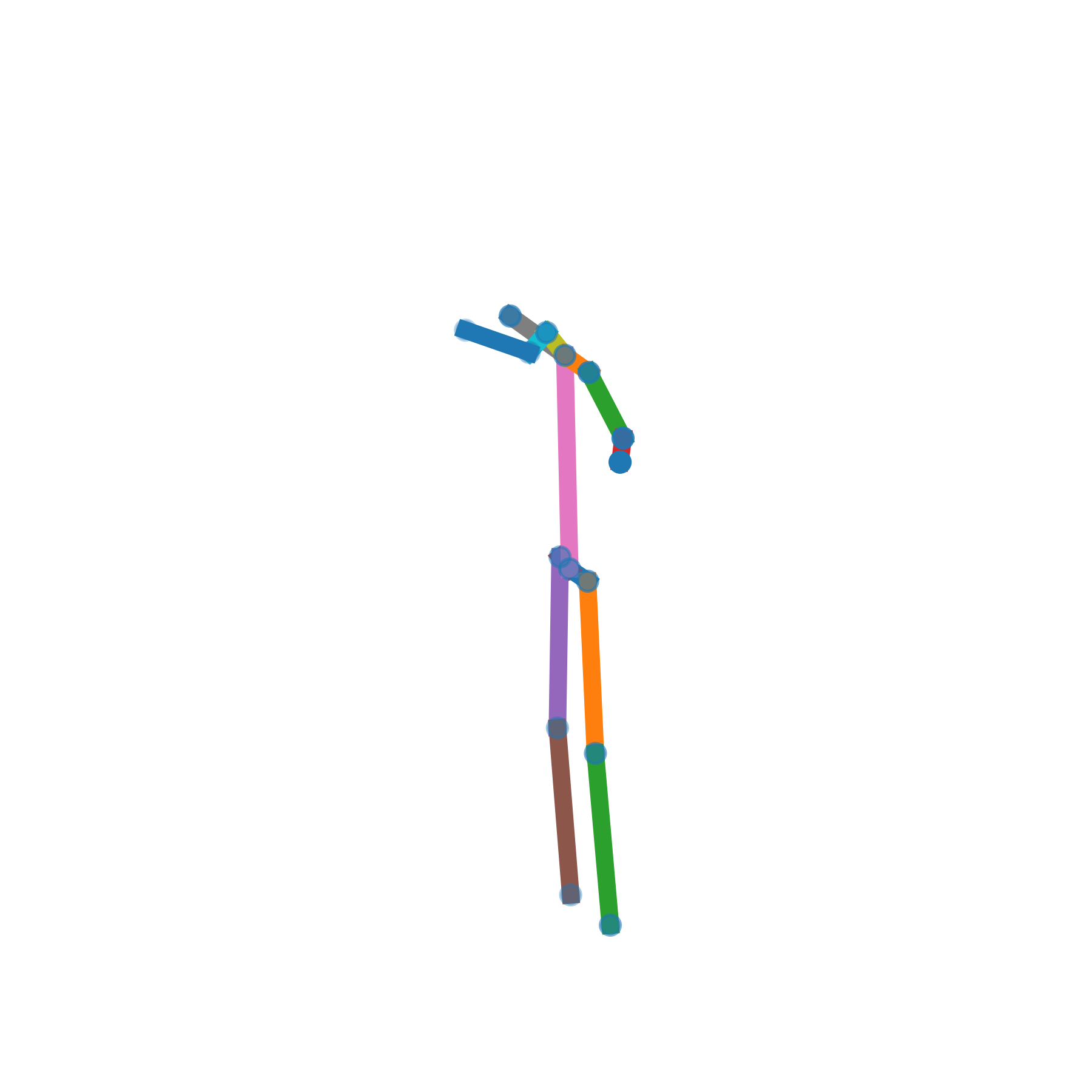}} \\[0.35cm]

        \multicolumn{5}{c}{\textbf{\textit{man maker} -- Frame 2007}} \\[0.1cm]

        \makebox[\linewidth][c]{\includegraphics[width=0.95\linewidth, trim=75 75 75 75, clip]{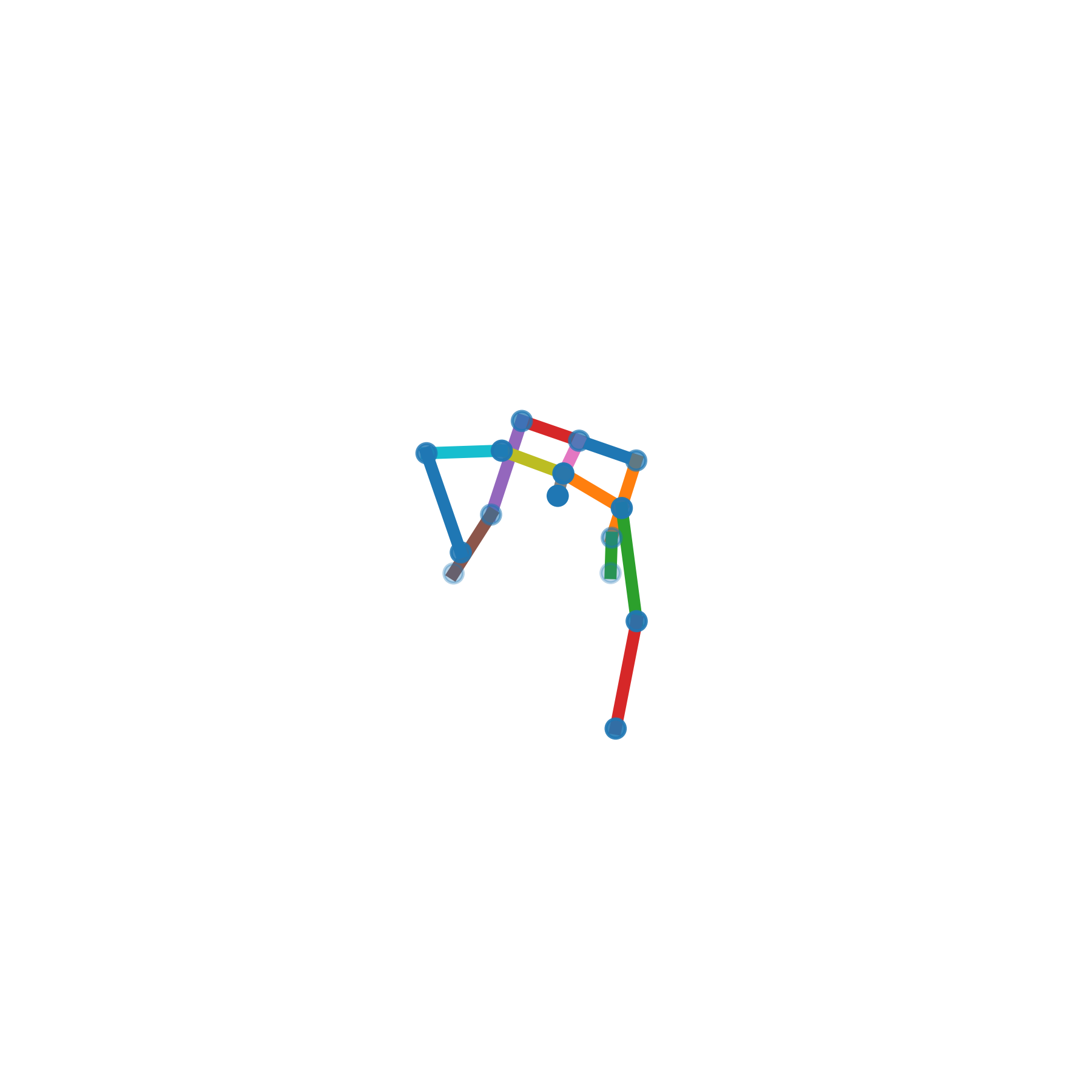}} &
        \makebox[\linewidth][c]{\includegraphics[width=0.95\linewidth, trim=75 75 75 75, clip]{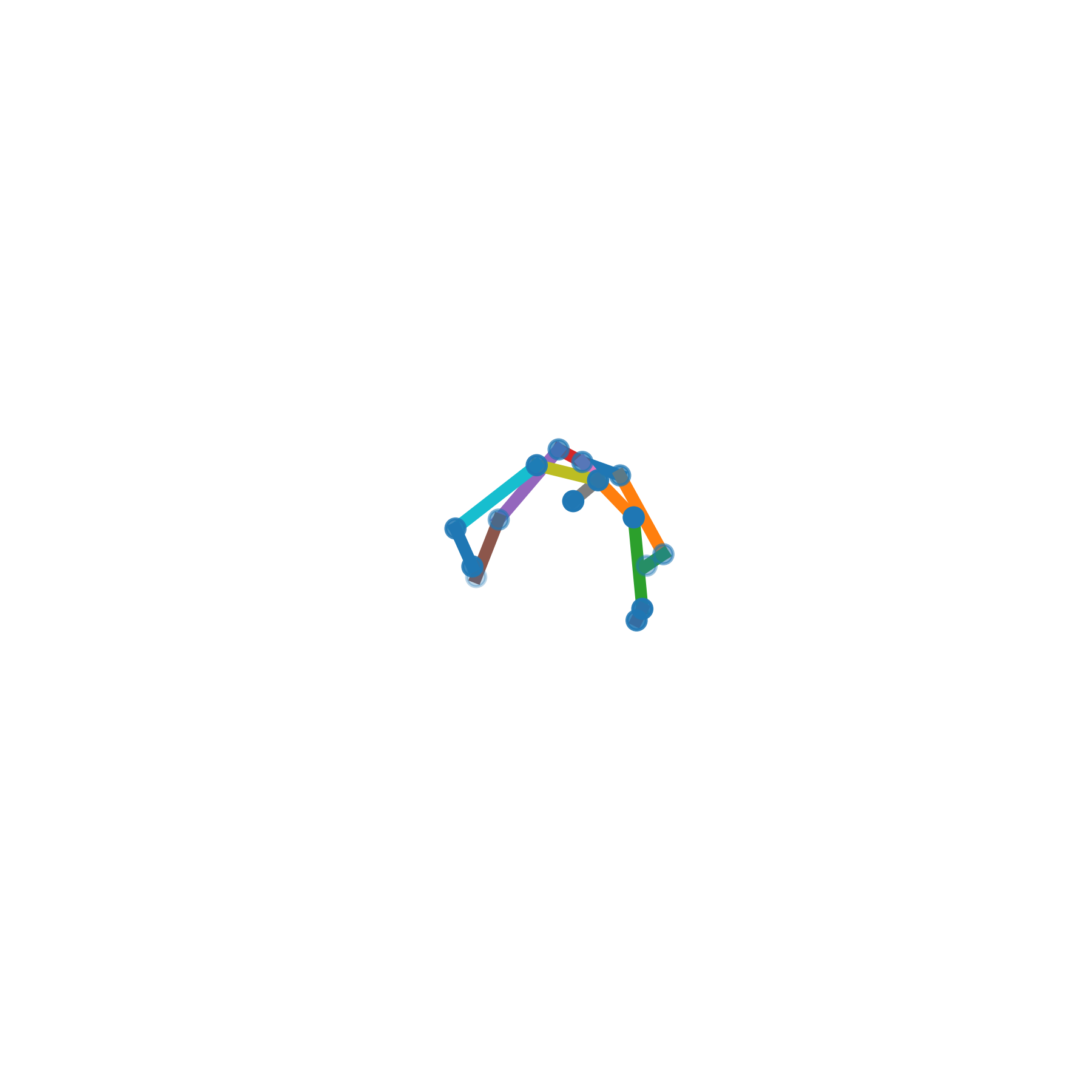}} &
        \makebox[\linewidth][c]{\includegraphics[width=0.95\linewidth, trim=75 75 75 75, clip]{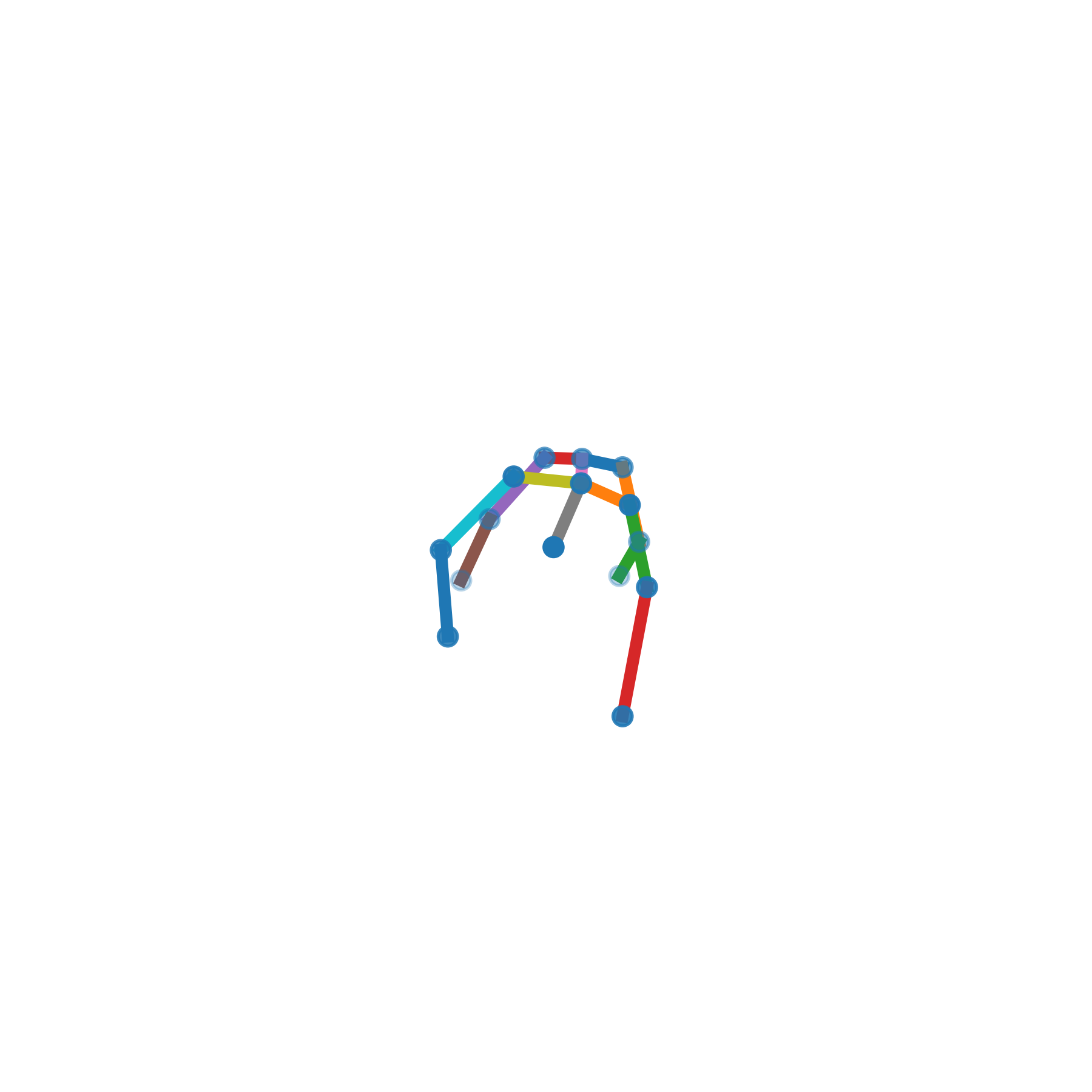}} &
        \makebox[\linewidth][c]{\includegraphics[width=0.95\linewidth, trim=75 75 75 75, clip]{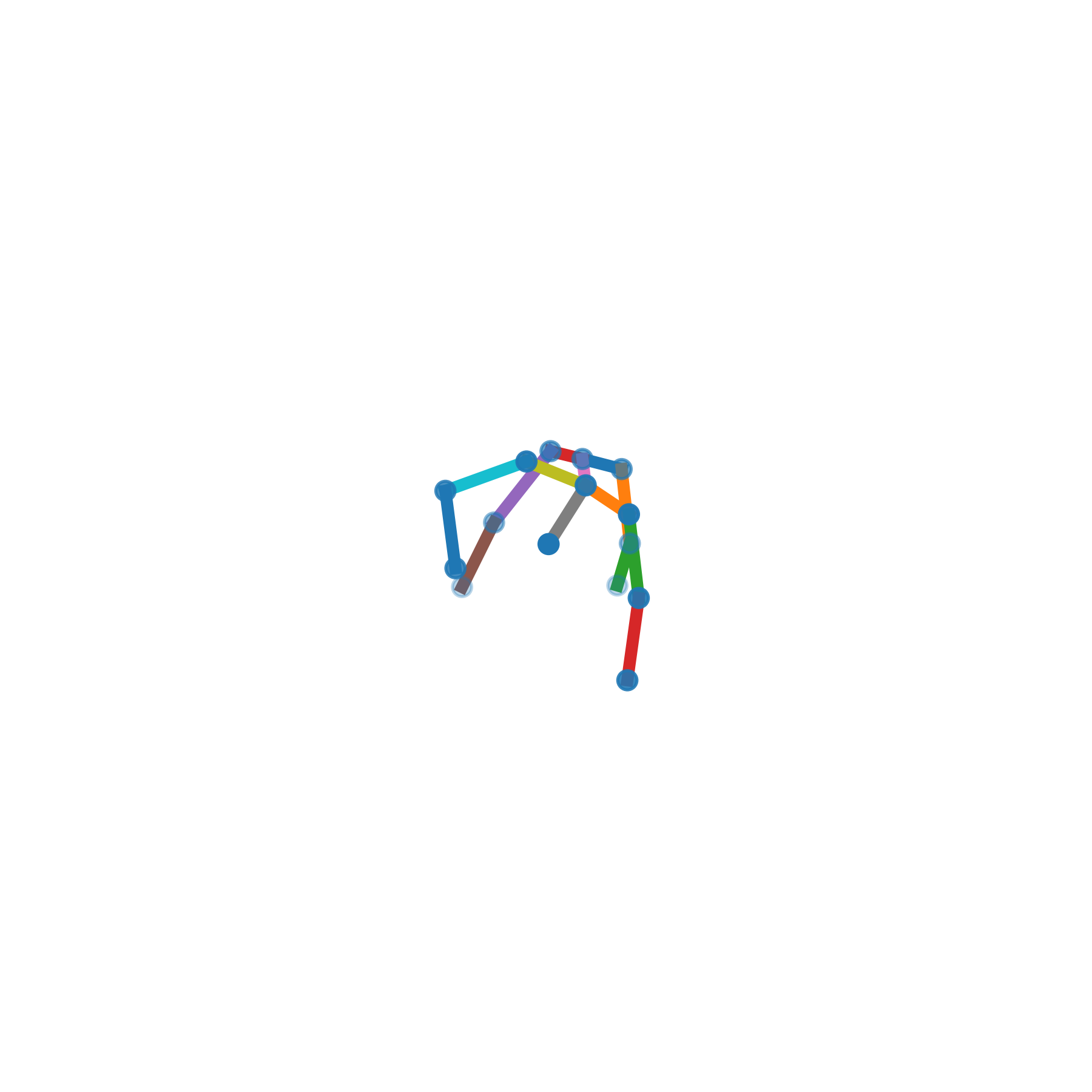}} &
        \makebox[\linewidth][c]{\includegraphics[width=0.95\linewidth, trim=75 75 75 75, clip]{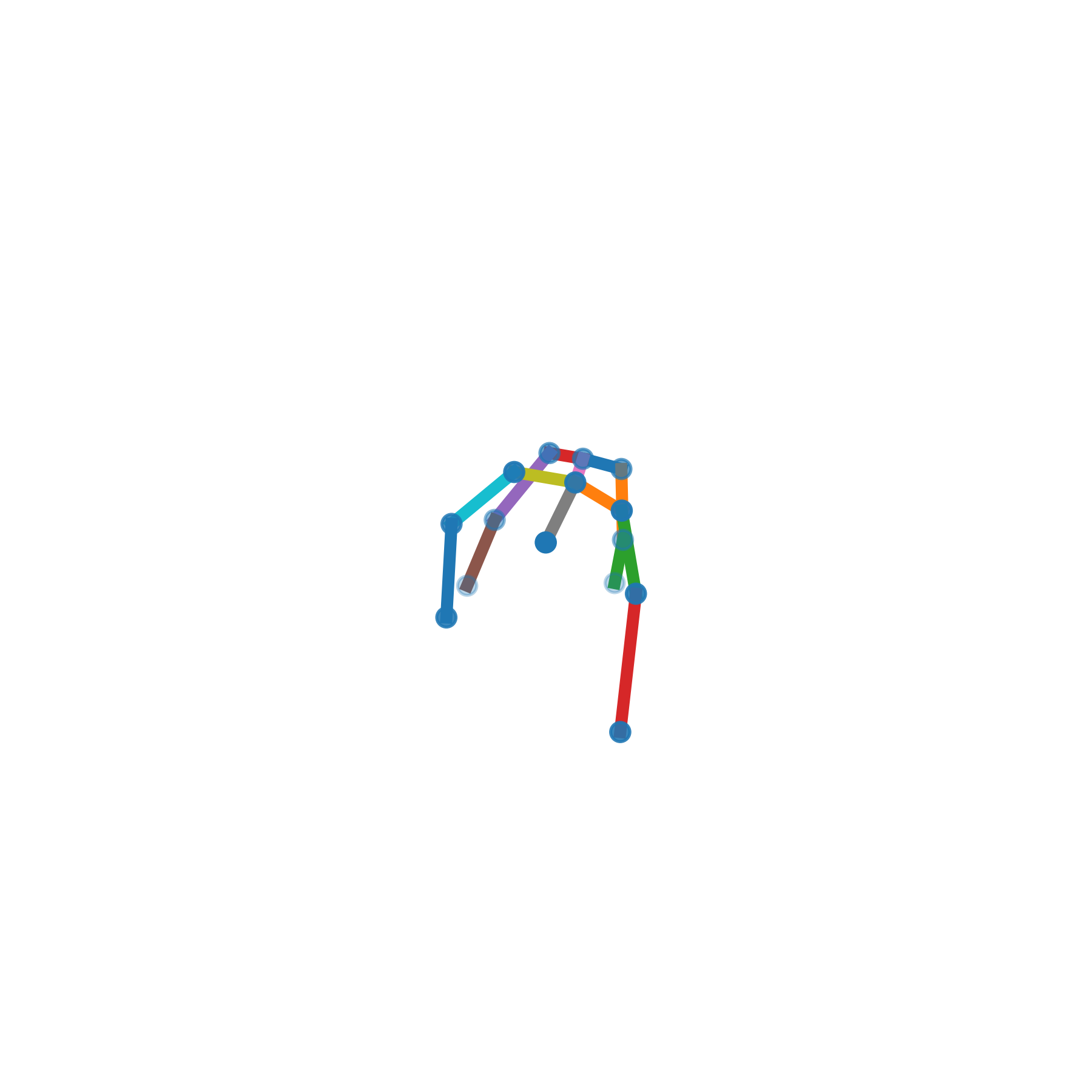}} \\

        \bottomrule
    \end{tabular}

    \caption{Qualitative comparison of MVGFormer predictions before and after temporal filtering in two representative sequences. The first row shows the \textit{band pull apart} sequence, characterized by a relatively regular standing motion, whereas the second row shows the more challenging \textit{man maker} sequence, which involves a complex floor-supported body configuration. Ground truth is compared with the original MVGFormer prediction and the outputs obtained after applying the One Euro, Savitzky--Golay, and Median filters. The examples illustrate how temporal filtering can attenuate isolated reconstruction errors and improve the consistency of the predicted pose, particularly in the Median-filtered output.}
    \label{fig:mvgformer_filtering_qualitative_comparison}
\end{figure}

Finally, Fig.~\ref{fig:mvgformer_filtering_qualitative_comparison} shows representative MVGFormer poses before and after filtering. The examples illustrate how temporal filtering reduces isolated reconstruction errors and improves pose consistency in both regular standing motion and more challenging body configurations.

\section{Discussion}
\label{se:discussion}

This study provides a unified evaluation of representative monocular and multi-view 3D human pose estimation methods under a common experimental protocol using the Fit3D dataset. Beyond comparing reconstruction accuracy, the experiments investigate how detector quality, temporal modeling, and post-processing influence the final performance of modern pose estimation pipelines. The results lead to several observations relevant to the design of practical pose estimation systems.

A key finding is the strong dependence of monocular lifting methods on the quality of the input 2D detections. Although this dependency is well known conceptually, our experiments quantify its magnitude using recent state-of-the-art architectures. Reconstruction errors for TCPFormer and PoseMamba-L increase noticeably when automatic detections are used, compared to the use of ground-truth 2D keypoints. Among the evaluated detectors, RTMPose consistently provides more reliable inputs than YOLO-Pose, resulting in lower reconstruction errors. These results show that improvements in 2D pose estimation translate directly into better 3D reconstruction accuracy, underscoring the importance of considering the full estimation pipeline rather than evaluating lifting models in isolation.

The comparison between TCPFormer and PoseMamba-L also shows that reconstruction performance cannot be explained solely by the average quality of the 2D detector. PoseMamba-L achieves lower errors with ground-truth 2D inputs, whereas with RTMPose detections both models obtain very similar average accuracy, with PA-MPJPE values of 70.89\,mm for TCPFormer and 71.14\,mm for PoseMamba-L. This close performance indicates that both temporal modeling strategies are competitive under realistic detection noise, while their relative behavior depends on the characteristics of the 2D input and on how each architecture handles the resulting temporal and spatial inaccuracies. When evaluated using COMMON15, the ranking is slightly reversed, although the difference between the two models remains below 1\,mm.

The comparison between monocular and multi-view approaches confirms the complementary strengths of both paradigms. MVGFormer achieves the lowest average spatial error in the COMMON15 evaluation, benefiting from synchronized observations from multiple calibrated cameras that reduce depth ambiguity and improve geometric consistency. Nevertheless, the advantage in spatial accuracy comes at the cost of a considerably more demanding acquisition setup, requiring multiple synchronized cameras, accurate calibration, and increased computational resources. In contrast, monocular lifting methods remain competitive despite relying on a single camera and externally detected 2D keypoints, making them attractive for applications where simplicity, portability, and deployment cost are primary considerations.

The temporal analysis reveals an equally important aspect that is often overlooked when evaluating 3D pose estimation systems. While MVGFormer achieves the lowest PA-MPJPE, its original predictions exhibit substantially larger velocity errors than the monocular models, reflecting the difference between frame-wise reconstruction and explicit temporal modeling. Consequently, spatial accuracy alone does not fully characterize the quality of reconstructed motion sequences, particularly in applications where velocities, accelerations, or movement smoothness constitute the quantities of interest.

The evaluation of temporal filtering shows that lightweight post-processing, which requires no retraining, can still be effective for modern deep learning models, although its benefit is strongly model-dependent. For the monocular models, Savitzky--Golay reduces MPJVE by approximately 2.34\% for TCPFormer and negligibly for PoseMamba-L, with practically unchanged spatial accuracy. For MVGFormer, the Median Filter reduces MPJVE from 26.46 to 4.19\,mm/frame and PA-MPJPE from 66.69 to 60.38\,mm. The most appropriate filter therefore depends on the temporal characteristics of the underlying model.

These results show that selecting an appropriate 3D human pose estimation pipeline requires balancing several competing objectives. Multi-view systems remain preferable when maximum spatial accuracy is required and suitable acquisition hardware is available. Conversely, modern monocular temporal models provide an attractive compromise between accuracy, computational efficiency, and ease of deployment, particularly when combined with accurate 2D pose detectors and lightweight temporal filtering. These findings provide practical guidelines for selecting pose estimation methods in applications ranging from sports performance analysis to rehabilitation and human motion understanding.

Although the proposed evaluation provides a comprehensive comparison of representative state-of-the-art methods, several limitations should be acknowledged. First, all experiments are conducted exclusively on the Fit3D dataset. Although Fit3D contains a diverse set of challenging fitness exercises, additional cross-dataset evaluations would be necessary to assess the generalization of the observed trends to other domains such as daily activities, clinical rehabilitation, or outdoor sports. Second, the evaluated models are used with publicly available pretrained weights rather than being fine-tuned on Fit3D. This decision enables a fair comparison of cross-dataset generalization, but it may not reflect the maximum performance attainable through dataset-specific optimization. 
Finally, the study focuses on three representative architectures; although these models cover the principal paradigms currently dominating the literature, future evaluations could incorporate recently proposed diffusion-based models, foundation models, or end-to-end image-based reconstruction methods.

\section{Conclusion}
\label{se:conclusion}

This work presented a unified evaluation of recent monocular and multi-view 3D human pose estimation methods on the Fit3D benchmark, analyzing the influence of the 2D detector, the temporal consistency of the reconstructed motion, and the effect of lightweight post-processing filters.

The main conclusion is that spatial accuracy alone is not sufficient to select a pose estimation pipeline for exercise analysis. The quality of the 2D detector is a major determinant of monocular lifting performance, while the reconstruction architecture also influences how detection errors are propagated to 3D. Multi-view reconstruction provides the best average spatial accuracy, although it requires a more complex acquisition setup and produces much less stable raw trajectories. Moreover, the most suitable temporal filter depends on the characteristics of the underlying model. In practice, a monocular temporal model combined with an accurate detector such as RTMPose and lightweight post-processing provides a competitive solution for single-camera applications, whereas MVGFormer benefits substantially from robust temporal filtering, with the Median Filter improving both temporal consistency and spatial accuracy in our experiments.

Future work will extend this evaluation to additional benchmarks and recently proposed architectures, including end-to-end image-based methods and foundation models for human motion understanding. We will also investigate domain adaptation techniques, fine-tuning strategies on fitness-oriented datasets, and task-specific evaluation protocols for biomechanics, rehabilitation, and sports performance analysis. We expect that the unified evaluation framework presented in this work will facilitate future comparative studies and contribute to a more comprehensive understanding of the strengths and limitations of modern 3D human pose estimation systems.

\section*{CRediT authorship contribution statement}
\textbf{N\'estor Ortega}: Conceptualization, Methodology, Software, Investigation, Data curation, Formal analysis, Writing -- original draft, Visualization. \textbf{Agust\'in Salgado}: Conceptualization, Supervision, Methodology, Writing -- review \& editing. \textbf{Javier S\'anchez}: Conceptualization, Supervision, Methodology, Writing -- review \& editing, Funding acquisition.

\section*{Acknowledgements}
This work was supported by the research project PID2024-155753OB-I00, funded by MICIU/AEI/10.13039/501100011033/FEDER, UE.

\section*{Data availability}
Fit3D~\cite{fit3d2024} is publicly available at \url{https://fit3d.imar.ro} under its academic license. The code used to preprocess the dataset, run inference with the evaluated models, and reproduce all reported metrics and figures, together with some of the 3D predictions used in the temporal smoothing experiments, is publicly available at \url{https://github.com/Nestpr/fit3d-3d-pose-benchmark} under the Apache~2.0 License.

\bibliographystyle{unsrt}  
\bibliography{references}  

\end{document}